%% file: main.tex
\documentclass[acmsmall]{acmart}

\AtBeginDocument{%
  }

\setcopyright{acmcopyright}
\acmJournal{PACMHCI}
\acmYear{2023} \acmVolume{7} \acmNumber{CSCW2}
\acmArticle{357} \acmMonth{10} \acmPrice{15.00}
\acmDOI{10.1145/3313831.3376873}

\usepackage{xspace}
\usepackage{subcaption}
\usepackage{booktabs}
\usepackage{enumitem}
\usepackage{svg}

\usepackage[normalem]{ulem}
\useunder{\uline}{\ul}{}

\usepackage{xcolor}
\usepackage{algorithm}
\usepackage{algpseudocode}

\newif{\ifhidecomments}
\ifhidecomments
    \newcommand{\chenhao}[1]{}
    \newcommand{\viv}[1]{}
    \newcommand{\vera}[1]{}
    \newcommand{\veraedit}[1]{}
     \newcommand{\chacha}[1]{}
\else
    \newcommand{\chenhao}[1]{\textcolor{blue}{[#1 ---\textsc{ct}]}}
    \newcommand{\viv}[1]{\textcolor{red}{[#1 ---\textsc{viv}]}}
    
    \newcommand{\chacha}[1]{\textcolor{magenta}{[#1 ---\textsc{chacha}]}}
    \newcommand{\vera}[1]{\textcolor{orange}{[#1 ---\textsc{vera}]}}
    \newcommand{\veraedit}[1]{\textcolor{orange}{#1}}
\fi

\newcommand{\para}[1]{\noindent{\bf #1}}
\newcommand{\figref}[1]{Figure~\ref{#1}}
\newcommand{\secref}[1]{Section~\ref{#1}}
\newcommand{\tabref}[1]{Table~\ref{#1}}

\newcommand{\imdb}{{\bf IMDb}}

\begin{document}

\title{Selective Explanations: Leveraging Human Input to Align Explainable AI}

\author{Vivian Lai}\authornote{denote equal contribution}
\email{vivian.lai@colorado.edu}
\affiliation{%
	\institution{University of Colorado Boulder}
	\city{Boulder}
	\state{CO}
	\country{USA}
}

\author{Yiming Zhang*}
\email{yimingz0@uchicago.edu}
\affiliation{%
	\institution{University of Chicago}
	\city{Chicago}
	\state{IL}
	\country{USA}
}

\author{Chacha Chen}
\email{chacha@uchicago.edu}
\affiliation{%
	\institution{University of Chicago}
	\city{Chicago}
	\state{IL}
	\country{USA}
}

\author{Q.
	Vera Liao}
\email{veraliao@microsoft.com}
\affiliation{%
	\institution{Microsoft Research}
	\city{Montreal}
	\country{Canada}
}

\author{Chenhao Tan}
\email{chenhao@uchicago.edu}
\affiliation{%
	\institution{University of Chicago}
	\city{Chicago}
	\state{IL}
	\country{USA}
}

\renewcommand{\shortauthors}{Lai et al.}

\begin{abstract}
    \input{abstract}
\end{abstract}



\maketitle

\input{sections/introduction}

\input{sections/related_work}
\input{sections/method}
\input{sections/study_1}

\input{sections/study_2}
\input{sections/discussion}

\begin{acks}
We thank the anonymous reviewers for their insightful comments.
This paper is supported in part by NSF grants, IIS-2040989 and IIS-2126602.
\end{acks}

\bibliographystyle{ACM-Reference-Format}
\bibliography{refs,yiming}

% For CSCW2 Articles V7cscw318-V7cscw373, use
\received{January 2023}
\received[revised]{April 2023}
\received[accepted]{May 2023}\textbf{}

\newpage
\appendix
\input{sections/appendix}

\end{document}

%% file: abstract.tex
While a vast collection of explainable AI (XAI) algorithms has been developed in recent years, they have been criticized for significant gaps with how humans produce and consume explanations. 
As a result, current XAI techniques are often found to be hard to use and lack effectiveness.
In this work, we attempt to close these gaps by making AI explanations \textit{selective}---a fundamental property of human explanations---by selectively presenting a subset of model reasoning based on what aligns with the recipient's preferences. We propose a general framework for generating selective explanations by leveraging human input on a small dataset. This framework opens up a rich design space that accounts for different selectivity goals, types of input, and more.
As a showcase, we use a decision-support task to explore selective explanations based on what the decision-maker would consider relevant to the decision task.
We conducted two experimental studies to examine three paradigms based on our proposed framework: in Study 1, we ask the participants to provide critique-based or open-ended input to generate selective explanations (self-input). In Study 2, we show the participants selective explanations based on input from a panel of similar users (annotator input). Our experiments demonstrate the promise of selective explanations in reducing over-reliance on AI and improving collaborative decision making and subjective perceptions of the AI system, but also paint a nuanced picture that attributes some of these positive effects to the opportunity to provide one's own input to augment AI explanations. Overall, our work proposes a novel XAI framework inspired by human communication behaviors and demonstrates its potential to encourage future work to make AI explanations more human-compatible.

%% file: sections/introduction.tex
\section{Introduction}
\label{sec:introduction}

With advances and widespread adoption of artificial intelligence (AI) systems, the need for people to understand AI in order to appropriately trust and effectively interact with AI has spurred great interest in the emergent field of explainable AI (XAI)~\cite{gunning2017explainable,gilpin2018explaining, doshi2017towards}. The technical field of XAI has made remarkable progress in recent years, producing a large collection of algorithms that aims to reveal the decision processes of machine learning (ML) models~\cite{guidotti2019survey,arrieta2020explainable}. 
However,
empirical human-subject studies that examine how people interact with state-of-the-art XAI techniques have not found conclusive evidence that these techniques help end-users better complete AI-assisted tasks~\cite{lai2021towards,bansal2021does,bussone2015role,buccinca2020proxy,green2019principles,lai2019human}. These AI explanations are often unintuitive and demand significant effort for people to process and understand~\cite{lage2019evaluation,friedler2019assessing,buccinca2021trust}. They have been found to risk impairing task performance~\cite{bansal2021does,green2019disparate,green2019principles,lai2019human,poursabzi2018manipulating}, efficiency~\cite{friedler2019assessing,lage2019evaluation,weerts2019human,abdul2020cogam,carton2020feature}, and user satisfaction~\cite{dietvorst2015algorithm,lucic2020does,lage2019evaluation}, ultimately preventing users from harnessing reals benefits due to explanations~\cite{buccinca2021trust,gajos2022people,vasconcelos2022explanations}. 

This difficulty to use current XAI techniques can be attributed to their lack of compatibility with how humans produce and consume explanations, as pointed out by social sciences literature~\cite{miller2019explanation,wang2019human,liao2021human}. For example, Malle's theory of explanation~\cite{malle2006mind} describes that a human explainer must engage in two fundamental processes to produce explanations---an process to \textit{gather} all reasons that can explain, and an impression management process to \textit{communicate} the explanation in social interactions. Arguably, by solely focusing on revealing the model decision processes, current XAI paradigms deal only with the \textit{reasoning} process and concern little with the \textit{communication} process.

How do people engage in explanation communication? 
Among other characteristics, people rarely present all explanatory causes but \textit{select} what they believe as serving the recipient's interest for achieving their goal, such as finding common grounds and providing new knowledge (more on selectivity goals to be discussed in \secref{rev:social-science}). 
That is, at the core of effective communication of explanation is \textit{explanation selection} based on the explainer's goal and their beliefs about the recipient. 

\begin{figure*}[t]
    \centering
    \includegraphics[width=0.95\textwidth]{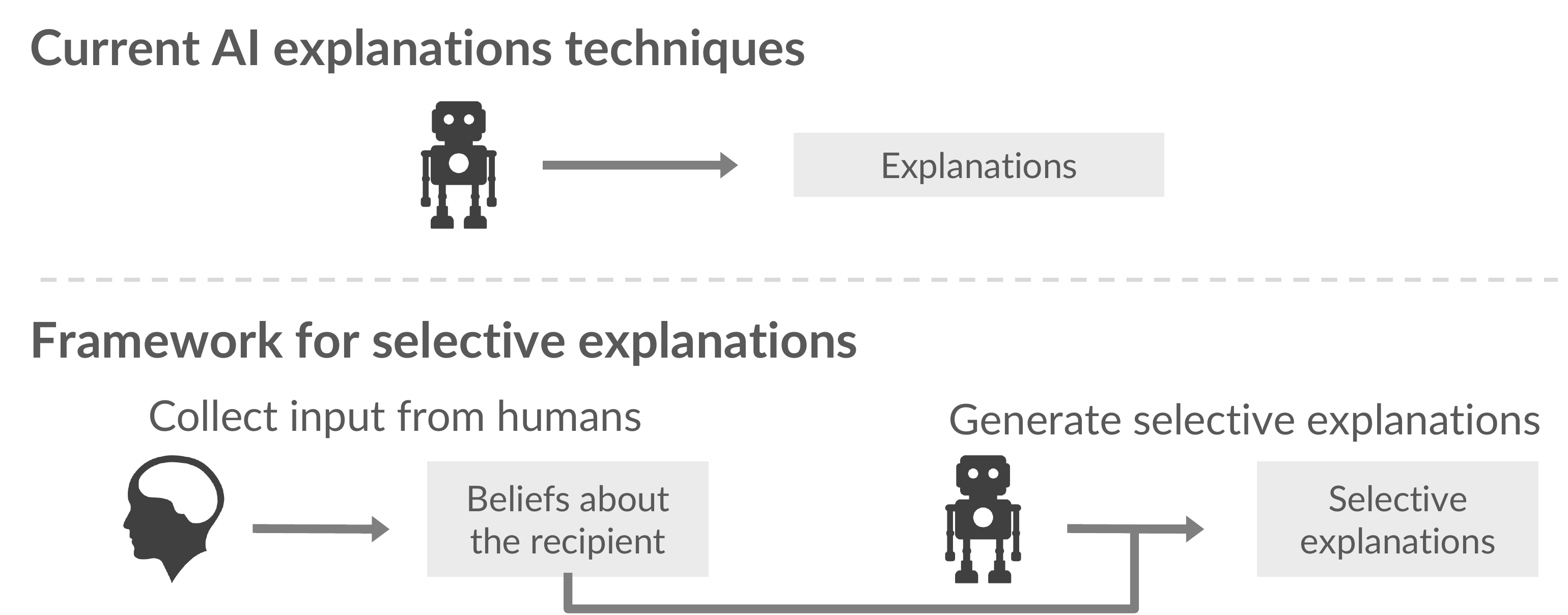}
    \caption{A high-level overview of the framework. See the full framework in  \figref{fig:design} and a detailed discussion in Section~\ref{sec:framework}.}
    \Description{screenshots of the interface}
    \label{fig:intro}
\end{figure*}

Inspired by this \textit{selective} property of human explanation, we introduce a novel framework to selectively present AI explanations based on beliefs about the recipient's preferences. This framework can be used to augment any existing feature-based local explanations---XAI techniques that explain a particular model prediction by how the model weighs different features of the instance, with potential of extending to other XAI techniques discussed in \secref{sec:discussion}. \figref{fig:intro} gives a high-level illustration of our framework (see the full version of our framework in \figref{fig:design} and a detailed discussion in \secref{sec:framework}).
On a high level, our framework consists of two steps: (1) collecting input from humans on a small sample and (2) generating selective explanations according to beliefs about the recipient's preferences for explanation, as inferred from the human input collected in step 1.

In \secref{sec:instantiation}, we present a way to instantiate this framework in a text classification task. Using this instantiation, we empirically explore the effects of selective explanations in an AI-assisted review sentiment judgment task as a testbed. Similar to people achieving better impression management with explanation selection, we expect selective explanations to improve users' perception of AI explanations. Meanwhile, a curious question is whether selective explanations can help people make better decisions.  Rather than ``enhancing trust'', the field of XAI is shifting its attention to ``calibrating trust''~\cite{zhang2020effect}, arguing that a more desirable goal of AI explanation should be to help people discern correct and incorrect model predictions to have more appropriate reliance on the model and thus more accurate human-AI joint decisions. We note that such a goal---facilitating the recipient to detect flaws of the explainer---is rarely the focus of human explanations, suggesting a possible tension with providing ``human-like explanations'' by AI. 

In \secref{sec:study_1} and~\ref{sec:study_2}, we explore these questions through two controlled human-subjects experiments (N=118, N=161) where we test three paradigms (from a broader space) of selective explanations based on our proposed framework. In Study 1, we ask participants to provide their own input to generate selective explanations (self-input), either with \textit{Open-ended} (selecting any features as aligning with their preferences) or \textit{Critique-based} (critiquing AI's explanation) feedback, and compare their effects to a baseline condition with non-selective explanations. In Study 2, we show participants selective explanations that are generated based on input \textit{from a panel of similar users} (annotator input). 

Results from these experiments demonstrate the promise of selective explanations. We found evidence that selective explanations were better aligned with the decision ground truth, improved decision outcomes, and decreased over-reliance when the AI predictions were wrong. They also consistently improved people's perceived understanding of the model over unselected explanations. Interestingly, in self-input paradigms, the opportunity to provide one's own input and have control over AI explanation also improved the perceived usefulness of AI, albeit at the cost of the increased overall workload. 

In summary, our main contributions can be summarized as follows: 
\begin{itemize}
    \item We propose a novel conceptual framework for generating selective explanations by leveraging human input and laying out the rich design space. Our work aligns with human-centered XAI efforts~\cite{liao2021human,vaughan2021humancentered,ehsan2021operationalizing} by providing a concrete way to operationalize human-like explanation communication behaviors that can be broadly applied to augment existing XAI techniques.
    \item We instantiate the framework in text classification and develop the corresponding algorithms and interface.
    \item We conduct two controlled experiments and demonstrate the promise of selective explanations in improving decision outcomes and subjective perceptions of AI.
\end{itemize}

In the rest of the paper, we first review related work that informed our research, then provide an overview of our framework. Then we instantiate the framework and present the two experiments exploring the effects of selective explanations. In \secref{sec:discussion}, we reflect on the results to discuss lessons learned, generalizability, and future directions, as well as open questions for our framework.

%% file: sections/related_work.tex
\section{Related Work}
\label{sec:related_work}

\subsection{Explainable AI and Its Pitfalls}
\label{sec:xai}
Recent years have seen a booming interest in explainable AI (XAI)~\cite{gunning2017explainable}, thanks to the unprecedented popularity of ``black-box'' AI models that are built on complex algorithms and architectures such as deep neural networks. Among the growing collection of XAI techniques (as surveyed in~\cite{adadi2018peeking,guidotti2019survey,gilpin2018explaining,carvalho2019machine,arrieta2020explainable}), we focus on those explaining deep machine learning classifiers (as opposed to other types of AI systems such as planning or multi-agent systems). ``Local'' XAI techniques that explain a model prediction (as opposed to ``global'' explanations to describe the entire model) can be roughly categorized into feature-based, example-based, and counterfactual explanations~\cite{guidotti2019survey}, with feature-based explanations being the most popular approach and the focus of this work. In short, feature-based explanations describe how the model weighs different features of the input instance to arrive at its prediction, often by highlighting the most salient features. As a general form of explanation, feature-based explanations can be generated by many different algorithms that vary in computational properties, such as LIME~\cite{ribeiro2016should} and SHAP~\cite{lundberg2017unified}.

Because explainable AI is fundamentally about supporting human understanding of models, the broad XAI community has been pushing for human-centered approaches~\cite{liao2021human,vaughan2021humancentered,ehsan2020human,ehsan2021operationalizing} that consider people's needs and preferences, as well as study how people actually interact with AI explanations. One line of such work focuses on summarizing common use cases of or objectives people have with AI explanations~\cite{arrieta2020explainable,liao2022connecting,suresh2021beyond,chen2022interpretable}, including supporting verifying and debugging models, assisting decision-making, auditing model (e.g., on bias, privacy and security issues), and knowledge discovery. Meanwhile, many HCI and CSCW researchers have explored developing XAI applications in various domains (e.g.~\cite{hohman2019gamut,xie2020chexplain,jacobs2021designing}), and conducting empirical studies to investigate the effects of explanations on people's task performance~\cite{zhang2020effect,bansal2021does,lai2019human}, efficiency~\cite{friedler2019assessing,lage2019evaluation,gero2020mental,weerts2019human,smithrenner2020,gonzalez2020human,yang2020visual,kocielnik2019will,lim2009and,cheng2019explaining,abdul2020cogam,levy2021assessing,carton2020feature}, cognitive load~\cite{abdul2020cogam,ghai2020explainable}, understanding~\cite{buccinca2020proxy,anik2021data,yang2020visual,lucic2020does,smithrenner2020,binns2018s,wang2021explanations,cheng2019explaining,cai2019effects}, subjective perceptions of AI~\cite{ghai2020explainable,biran2017human,tsai2021exploring,dietvorst2018overcoming,kocielnik2019will,lucic2020does,lage2019evaluation,narayanan2018humans,cai2019human} among others.

Unfortunately, results from these recent empirical studies of AI explanations are mixed at best. On the one hand, many studies found positive evidence that explanations improve people's understanding of the model~\cite{cheng2019explaining,ribeiro2018anchors,lakkaraju2016interpretable,lim2009and}, enhance people's subjective perception of and tendency to follow AI~\cite{panigutti2022understanding}, help data scientists debug the model~\cite{hohman2019gamut,narkar2021model}, and auditors detect model biases~\cite{dodge2019explaining}. On the other hand, multiple studies reported that end-users found the explanations generated from popular technical approaches hard to use, distracting, time-consuming, and cognitively demanding~\cite{jacobs2021designing,xie2020chexplain,springer2019progressive,liao2020questioning,robertson2021wait}. Due to the added cognitive load, studies also found that showing explanations reduce task satisfaction for people with a low ``need for cognition'' trait~\cite{ghai2020explainable,buccinca2021trust} (not enjoying cognitively demanding activities). These surprisingly negative effects of explanations from empirical studies have been referred to as XAI pitfalls~\cite{liao2021human,ehsan2021explainability}

In particular, recent studies begin to call out a prominent XAI pitfall---increasing people's over-reliance when the AI is wrong, which is especially problematic in the common use case of XAI for decision support. While the expectation is that explanations can  help people detect flawed model reasoning and make better decisions, empirical studies either failed to observe this effect~\cite{wang2021explanations} or even found the opposite that explanations make people more likely to blindly follow the model when it is wrong compared to showing only AI predictions~\cite{zhang2020effect,bansal2021does,poursabzi2021manipulating,wang2021explanations}. Research has attributed this phenomenon to a lack of cognitive engagement with AI explanations~\cite{kaur2020interpreting,buccinca2020proxy,gajos2022people,liao2021human}: when people lack either the motivation or ability to carefully analyze and reason about explanations, they make a heuristic judgment, which tends to superficially associate being explainable to being trustworthy~\cite{liao2022designing,ehsan2021explainable}. A recent CSCW work by~\citet{vasconcelos2022explanations} further calls out that this lack of cognitive engagement will persist if XAI techniques remain hard to use, as people strategically choose between engaging with explanations and simply deferring to AI after weighing the cognitive costs.

Motivated by these prior works, we aim to make AI explanations more human-compatible by making them easier to use and thereby tackling these XAI pitfalls. To explore the benefits of the proposed approach, as informed by prior empirical studies of XAI for decision-support, we will measure participants' decision outcomes, reliance on AI, efficiency, subjective cognitive load,  understanding and perceived usefulness of the AI. 

\subsection{Making Explainable AI Human-Compatible: The Case for Selectivity}
\label{rev:social-science}
While HCI researchers have taken various efforts to design XAI systems that are more user-friendly~\cite{jacobs2021designing,xie2020chexplain}, less cognitively demanding~\cite{abdul2020cogam}, or nudge people to better engage with explanations~\cite{buccinca2021trust},
current XAI techniques' difficulty to use can be fundamentally attributed to their disconnect with how humans produce and consume explanations~\cite{miller2019explanation,wang2019designing,liao2021human}. Such criticism is best reflected in Miller's work~\cite{miller2019explanation} that brings insights about human explanations from social sciences literature and argues that XAI should be built with human explanation properties in mind. Miller summarizes three fundamental properties of human explanations: contrastive (against counterfactual scenarios), selective, and social (as part of social interactions). This work has since inspired many new techniques aiming to make AI explanations more human-compatible, such as counterfactual~\cite{verma2020counterfactual,wachter2017counterfactual} and weight-of-evidence explanations~\cite{alvarez2021human} that cater to the contrastive property, and various kinds of interactive explanations~\cite{slack2022talktomodel,zhang2022towards} inspired by the social property. 

Our work is directly inspired by the selective property, which Miller points out as missing from current XAI techniques. As discussed in \secref{sec:introduction}, human explanations are often selected for social and cognitive reasons~\cite{hilton2017social,malle2006mind}, as the complete reasoning or causal chains are often too large to comprehend (e.g. the causes of a fatal car accident can be explained by a chain of a few dozen of events). There has been a line of psychology work arguing explanation selection is not arbitrary but follows common criteria~\cite{hesslow1988problem}. For example, \citet{hilton1986knowledge} demonstrate that abnormal (unusual or rare events) factors or events are more often presented in explanations while commonplace knowledge is often omitted (e.g., an unexpected lane change versus driving at 75 mph on a highway). ~\citet{hilton2007course} show that intentional actions that are deliberately changed (e.g., the driver is drunk), and relatedly, controllable events~\cite{mcclure1997you} that can be changed with intentions, tend to take priority in explanations. Many also suggest that people prioritize the most important or relevant reasons, which can be matters of necessity, sufficiency, or robustness in causal reasoning~\cite{woodward2006sensitive,lipton1990contrastive}. Our proposed framework is directly inspired by this body of literature on how humans selectively present explanations, with some of of the most common criteria being relevance~\cite{woodward2006sensitive,lipton1990contrastive}, abnormality~\cite{hilton1986knowledge}, and changeability~\cite{hilton2007course,mcclure1997you}.

\subsection{Learning from Human Input on Explanations} 
Our work is also informed by a small but growing set of works on eliciting human input relevant to AI explanations. Prior work explored eliciting human feedback on model explanations as additional supervision signals for model training~\cite{ghai2020explainable,stumpf2009interacting,cartonWhat2022}. For example, \citet{ghai2020explainable} proposed explainable active learning where labelers are asked to not only provide labels to train the model but also critique feature-based explanations produced by the learning model. Other works proposed new XAI techniques by eliciting human's own rationales (e.g. which keywords are important or what rules to follow to reach decisions)~\cite{ehsan2019automated} or domain concepts~\cite{kim2018interpretability} to help generate or improve AI explanations. For example, \citet{ehsan2019automated} propose to train an explanation generation model directly from elicited human rationale data to help lay users make sense of model actions. Another relevant work by \citet{fenglearning} trains a model to select different combination of explantions to accommodate different users' needs and preferences. However, we note that ``selectivity'' in this paper is about selection from multiple explanation sources using user feedback, rather than the selectivity demonstrated in human explanation communication, which is the focus of our work.

Instead of proposing a new XAI algorithm, we propose a novel framework that can be broadly applied to \textit{augment} the outputs generated by any existing feature-based XAI algorithm. Closest to our work is a recent study by~\citet{boggust2022shared}. To help people better and more efficiently analyze model behaviors, they propose a set of metrics to contrast model reasoning via the saliency method (a feature-based explanation for image data) and human reasoning gathered from annotations. To name a few, the ``human aligned'' metric measures how often human and model reasoning are consistent, and the ``sufficient subset'' metric measures the degree to which model rationale contains human rationale. These metrics can then be used to rank and sort a large number of data instances, helping people identify and analyze different patterns of model behavior. Our work is inspired by this general approach of contrasting raw outputs generated by XAI algorithms and human rationale, but we leverage the latter to augment the former and lay out the design space by considering different goals, as well as ways to elicit human rationale, and present selective explanations.
To measure the effectiveness of our proposed framework, we investigate the effect of selective explanations through controlled human-subject experiments. 

%% file: sections/method.tex
\section{Framework: Selective Explanation with Human Input}
\label{sec:framework}

Inspired by the selective property of human explanation, we propose a general framework for generating selective AI explanations by leveraging human input. Our framework consists of two steps: the input step and the selection step.

In the \textbf{input step}, the goal is to elicit human input that  can be used to infer beliefs about the user, such as which features the user would consider relevant to the decision task. Ideally, this step should be efficient and requires only a small sample set to elicit human input.

In the \textbf{selection step}, a separate prediction model, which we call \textit{belief prediction model} hereafter, is used to generalize from input gathered in the first step to predict the recipient's preferences regarding the given instance of explanation, then selectively augment the original explanation by prioritizing features that align with the predicted preference. At the heart of our framework is an algorithm of this belief prediction model, for which we propose a simple yet effective approach in \secref{sec:instantiation}.

\begin{figure*}[t]
    \centering
    \includegraphics[width=0.95\textwidth]{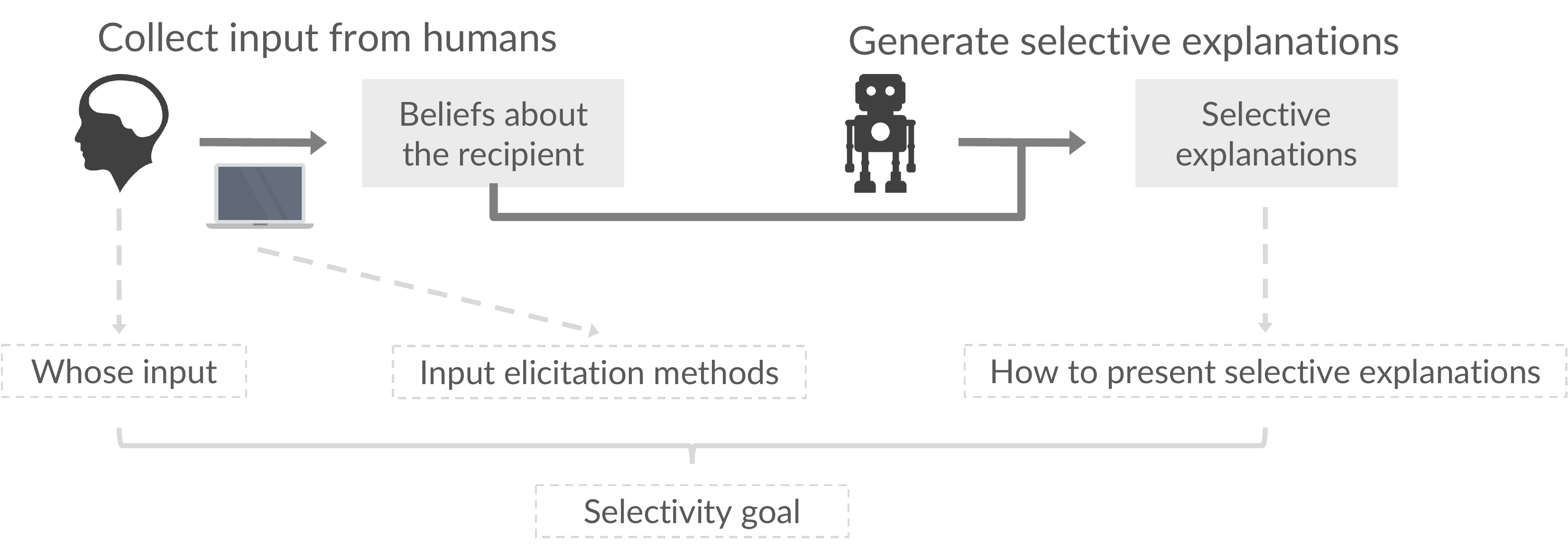}
    \caption{Illustration of the design space in our framework. The dashed-line boxes highlight dimensions of design considerations.}
    \Description{Illustration of the design space in our framework. The dashed-line boxes highlight the design considerations.}
    \label{fig:design}
\end{figure*}

We start by discussing the design choices under our framework (see \figref{fig:design} for an illustration), and create an instantiation in \secref{sec:instantiation} using a subset of this design space to conduct empirical studies.
While we limit our user studies to local feature-based explanations, we will discuss algorithmic considerations for generalizing beyond this particular instantiation and feature-based explanations in \secref{sec:discussion}.

Our entire framework is contingent on the \textit{selectivity goal} of the model, which may vary across XAI use cases. In the input step, the input can be obtained from \textit{different kinds of stakeholder group} with different \textit{elicitation methods}. In the selection step, once the selective explanations are generated, the key question is \textit{how to present} them visually. Below we elaborate on these dimensions of design choices. Our goal is to explore this design space and layout possibilities for future work to utilize this framework, rather than to make conclusive recommendations. In \tabref{tab:framework_overview}, we list examples of how to make choices in these design dimensions (excluding the \textit{how to present} dimension) for popular XAI use cases discussed in the literature~\cite{arrieta2020explainable,liao2022connecting,suresh2021beyond,chen2022interpretable}. We will refer to this table throughout the discussions below.

\(\bullet\) \textbf{Selectivity goal}: Drawing from social science literature on common criteria based on which humans produce selected explanations (reviewed in \secref{rev:social-science}), we suggest three general selectivity goals that can appear in different XAI applications: relevance, abnormality, and changeability---future work can further expand these goals. In \tabref{tab:framework_overview}, we list examples of XAI use cases where each selectivity goal is appropriate.
    
- \textit{Relevance} prioritizes reasons that the recipient would deem relevant or important to the task. In different XAI use cases, relevance may have specific meanings. For example, for a decision-support AI that helps people detect review sentiment, the relevance goal would prioritize presenting features (i.e., words) relevant for judging the sentiment. When applying XAI for auditing model biases, relevance would focus on features related to protected attributes such as race and gender (including potentially correlated features, e.g., zip code). By producing explanations that are more concise and relevant to what the recipient is looking for, we hypothesize that the relevance goal can help the recipient discover useful information more easily, improve the intuitiveness, ease of use, and overall perception of explanation.

- \textit{Abnormality} prioritizes reasons that the recipient would find abnormal or surprising. For example, when applying explanations to debug why the model makes certain mistakes, the abnormality goal could highlight features that the model unexpectedly (for the person doing debugging) picked up in its decision process to help people detect model abnormality more accurately and efficiently. In XAI use cases for knowledge discovery (e.g., supporting data analysts), if the model process is verifiably correct, this abnormality goal could be used to help people learn new knowledge such as identifying contributing factors that are unknown to the user. Note that in some use cases (e.g., learning new knowledge about judging review sentiment), the abnormality goal can be seen as the reverse of the relevance goal---while the latter selectively prioritizes reasons that align with the recipient's rationale, the former prioritizes reasons that do not align with human intuition but are nevertheless useful.  

- \textit{Changeablity} prioritizes reasons that can be changed or are more easily changeable. This goal is especially helpful for XAI use cases where explanations are sought for recourse~\cite{karimi2021algorithmic}---taking actions that can result in a different, often more desirable prediction in the future. For example, if an applicant's loan application is rejected due to an algorithmic risk assessment tool, an explanation should prioritize features that they can take action to change (e.g., reducing frequency of credit inquiry) and de-emphasize what they cannot easily change (e.g., significantly increasing income). While counterfactual explanations~\cite{wachter2017counterfactual,verma2020counterfactual}, which automatically search for features that with minimum change can alter the prediction, are often proposed to support recourse, existing techniques do not consider the changeability of the features shown and thus have been criticized for lacking actionability~\cite{barocas2020hidden,verma2020counterfactual}.

\(\bullet\) \textbf{Whose input:} Another key design dimension under this framework is from whom to elicit the input in the first step. In a most straightforward form, the input can come from the \textit{individual recipient} who will receive the explanations (\tabref{tab:framework_overview} gives specific examples of who the individual recipient is according to the XAI use case). However, this approach creates additional workload and requires time and resources that not every individual can afford.  Alternatively, one may assume there is shared preferences for a task, and collect human input via \textit{a panel of annotators similar to the target users} and apply their input to generative selective explanations for all. In some situations, individuals may lack the domain knowledge to effectively articulate what is relevant or abnormal. One may choose to gather input from an "ideal user archetype", such as \textit{domain experts}, and use the input to improve the experience for all. Importantly, different choices of ``whose input'' can introduce different effects and even biases, and must be carefully tested and justified for a specific XAI use case. We will empirically explore the differences between eliciting input from individual recipients versus a panel of annotators, and further reflect on this design dimension in \secref{sec:discussion}.

\(\bullet\) \textbf{Input elicitation method:} Once the selectivity goal is determined for a specific XAI use case, the elicitation asks the human input provider ``which features should be considered as relevant/abnormal/changeable for this use case? ''  While it is possible to ask such a question in the absence of context, the knowledge elicitation literature~\cite{cooke1994varieties} suggests that people are often better at articulating their knowledge or opinions with examples. Therefore, we suggest eliciting human input based on a small sample of examples. Example-based input can be \textit{open-ended}---asking directly to pick features from the example, or \textit{critique-based}---asking for agreement or disagreement with AI explanations for the given example. In \tabref{tab:framework_overview}, we list example questions to ask for specific XAI use cases and selectivity goals, focusing on the open-ended feedback (critique-based feedback would simply require pointing to the model explanation, e.g., ``which features \textit{in the model explanation} are relevant'' ). We generally recommend lower-precision input as natural human rationales are often qualitative~\cite{miller2019explanation}. That is, the elicitation could ask the person to select relevant features or rank feature by their relevance, as opposed to specifying precisely how relevant each feature is.

A contingent design decision here is the \textit{sampling strategy} to select examples to get the input. While a simple strategy could be random sampling or sampling examples with high-coverage features (i.e., shared by many instances), there exist more sophisticated strategies that depend on the selectivity goal (e.g., searching for examples with possible abnormalities). Furthermore, many design decisions can be made at the elicitation interface level, such as the modality (e.g., graphic vs. conversational interface) and language styles.

\(\bullet\) \textbf{How to present selective explanation:} In the last step, once the selective explanation can be generated, one needs to decide how to present it visually. This decision depends on UI characteristics, user preferences, and the holistic system user experience. We can only propose a few possibilities. We start by considering popular UI designs for non-selected feature-based explanations: for text or image data, saliency map is often used to visually highlight important keywords or superpixels (perceptual grouping of pixels). For tabular data, a horizontal bar chart is often used to visualize the importance of different features in the given instance. One possibility is to only present features that align with the predicted recipient preferences and hide presenting misaligned features. While this approach can produce explanations that are the most lightweight visually, it comes with a \textit{tradeoff of faithfulness}---losing information about how the model actually works. To mitigate this tradeoff, an alternative is to still maintain the presence of misaligned features, but augment them with different visual cues, such as by graying out or using an underlying waveline. For cases where faithfulness is critical (e.g., debugging~\cite{rudin2019stop}), one can preserve the original explanations and add additional highlights to the aligned parts.

\begin{table*}
\sffamily
\centering
\begin{tabular}{p{0.22\textwidth}p{0.22\textwidth}p{0.18\textwidth}p{0.26\textwidth}} 
 \toprule
\textbf{Example XAI Use Case} & \textbf{Selectivity Goal and Benefit} & \textbf{Whose Input (individual recipient)} & \textbf{Example Questions for Input Elicitation (open-ended)}\\
 \toprule
 AI assisting consumers to detect review sentiment & Relevance: improve ease of use & Consumers & Which words are relevant for judging the example review's sentiment? \\
 \midrule
 AI assisting loan officers to assess loan application risk & Relevance: improve ease of use & Loan officers & Rank the features by their importance for assessing the example applicant's risk. \\
\midrule

Audit model biases in recidivism prediction & Relevance: improve ease of use & Auditors & What features are relevant for making unfair predictions (e.g. protected attributes)?\\
\midrule

 Debug classification models & Abnormality: help detect model errors accurately and efficiently& Machine learning engineers & Which features should the model NOT base its decisions on?  \\
 \midrule

 Assist knowledge discovery for sales analysts & Abnormality: help detect unknown patterns efficiently & Analysts & Which features are less familiar for you to know how they may predict the sales outcomes?\\
 \midrule

Support recourse for loan applicants & Changeability: facilitate actionable changes & Loan applicants & Which features are possible/require less effort for you to make changes on? \\
 \bottomrule
\end{tabular}
\caption{Illustration of design choices made with our framework for common XAI use cases. All the columns should be taken as examples instead of best practices.
}
\label{tab:framework_overview}
\end{table*}

\section{Instantiating the Framework: Predicting and Explaining Movie Review Sentiment}
\label{sec:instantiation}

Building on our proposed framework, we develop a testbed in the context of AI-supported sentiment judgment. We first discuss the task, model, and base explanations, then the decision choices we made for each design dimension of our proposed framework and how selective explanations are generated. We will use this instantiation to conduct two empirical studies described in \secref{sec:study_1} and \ref{sec:study_2}.

\paragraph{Task, model, and explanations}
We choose a sentiment analysis model as our testbed because it is one of the most studied problems in classification~\citep{pang2008opinion}.
In addition, prior work has shown that explanations can increase over-reliance on AI when it is wrong even in this relatively simple task to humans~\citep{bansal2021does}.

We train a movie sentiment prediction model using a dataset of IMDb movie reviews (\imdb)~\citep{maasLearningWordVectors2011}.
\citet{maasLearningWordVectors2011} collected a balanced set of 50,000 reviews, where negative reviews have scores $\le 4$ and positive reviews have scores $\ge 6$. We randomly sampled without replacement to obtain three subsets: a training set of 200 examples, a development set of 500 examples, and a test set of 500 examples. Because sentiment analysis is a relatively easy classification task (state-of-the-art models can achieve an accuracy of almost 95\%~\citep{yangXLNetGeneralizedAutoregressive2020}), we intentionally used a small training set so that the model would perform less than perfectly. This set-up would require people to make more careful judgments with each AI prediction and allow us to study human-AI collaborative decision-making. Specifically, we use a BERT~\citep{devlin2018bert} model (\textbf{bert-base-uncased}) as the backbone architecture. Following the standard practice, we fine-tune a linear layer on top of the language model with a learning rate of $5 \times 10^{-5}$ and a batch size of 128 for 200 steps. The fine-tuned model achieves an accuracy of 85.2\% on the IMDb test set.

\begin{figure*}[t]
    \includegraphics[width=0.95\textwidth]{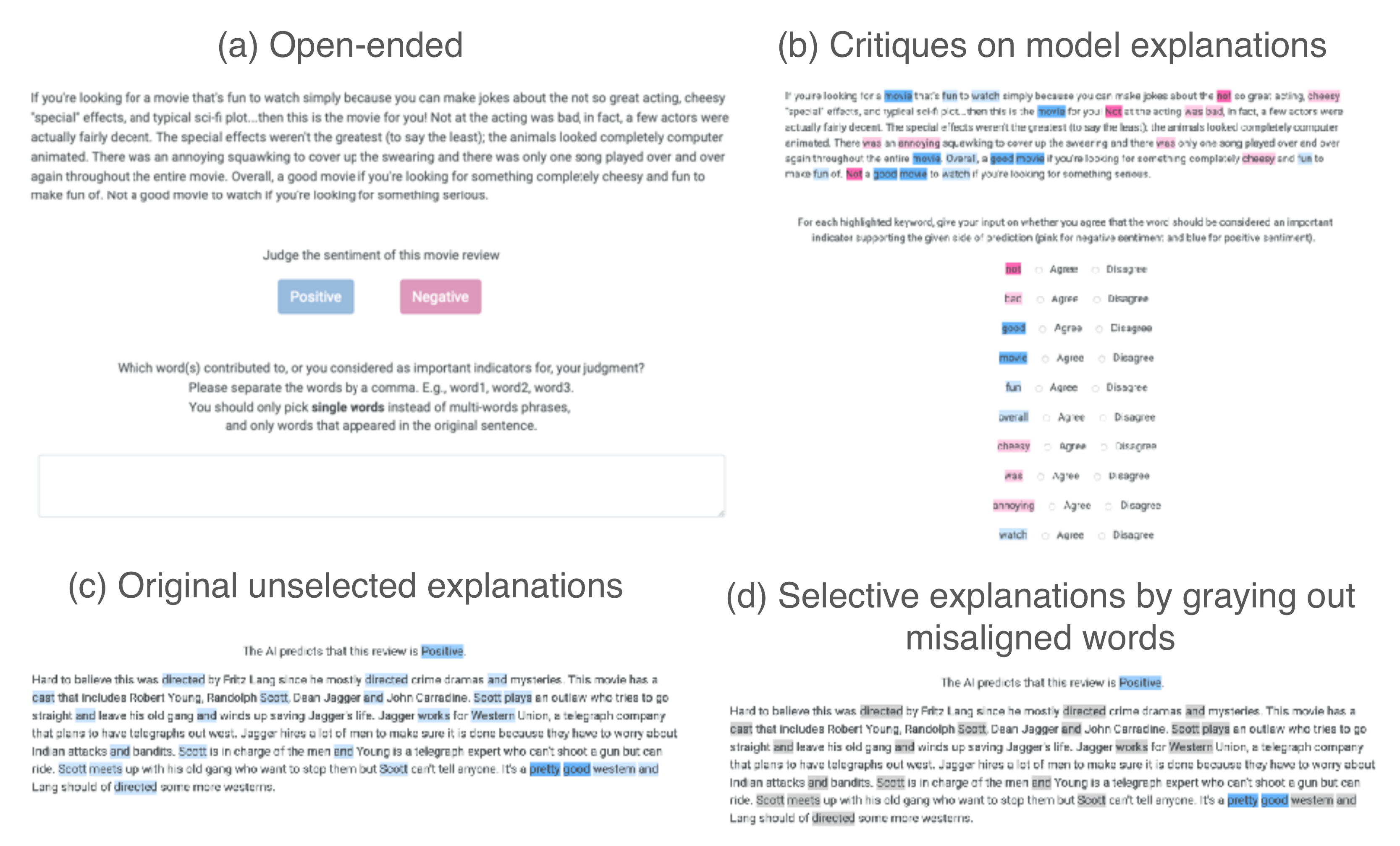}
    \caption{Instantiation of the framework in this work. We consider two approaches to collecting human inputs: (a) open-ended and (b) critique-based, and present (d) selective explanations by graying out features that are predicted to be misaligned with what the user would consider as relevant for judging review sentiment. (c) represents the original explanations generated by LIME.}
    \Description{screenshots of the interface}
    \label{fig:instantiation}
\end{figure*}

We use feature-based explanations by highlighting important words contributing to the model prediction for text classification. We apply LIME~\citep{ribeiro2016should}, a popular post-hoc XAI algorithm, to generate the importance scores for each word---measuring the degree to which it contributes (positively or negatively) to the model's prediction. LIME estimates this importance score by fitting a sparse linear bag-of-words model to locally approximate the BERT model. Then, we take the unique words with top-10 importance scores as the keywords explanation set for \textit{why} the instance gets a particular prediction (every occurrence of a word is highlighted). Note this explanation set could include both positive and negative keywords. For example, a movie review may have 8 keywords with positive weights for positive sentiment, and 2 keywords with negative weights. The fact that the majority of keywords are positive explains why the review is predicted to be positive. We visually present the explanations with saliency highlights: as shown in \figref{fig:instantiation}, we highlight the keywords, with colors indicating the direction of the weights (blue for positive sentiment, red for negative sentiment), and shades indicating the importance of the features (e.g., dark red means the word strongly contribute to a prediction of negative sentiment).

\paragraph{Design choices in instantiating the framework of selective explanations} We make the following design choices to study in our empirical studies out of a larger possible set of choices based on our proposed framework. We will reflect on these choices and discuss alternatives in \secref{sec:discussion}.

\(\bullet\) \textbf{Selectivity goal: } We focus on the goal of \textit{relevance} for an AI-assisted decision-making task since the explanations are expected to help the decision-makers discover relevant information and should be intuitive to use. That is, the selective explanation should prioritize presenting words (from its original explanation) that the recipient would consider relevant for judging movie sentiment. 

\(\bullet\) \textbf{Whose input: } We choose to study two possible scenarios to empirically explore the effects of different design choices in this dimension.
In Study 1, we ask each \textit{individual user} to provide input and the selective explanations are thus personalized.
In Study 2, we obtain input from a \textit{panel of similar users} so the selective explanation is fixed for all participants for a given input. 

\(\bullet\) \textbf{Input elicitation methods:} We choose two elicitation methods to be compared in Study 1: \textit{open-ended} and \textit{critique-based} input. Specifically, for the input phase, we present a sample of movie reviews to people and ask them to provide input for each review. For open-ended input, as shown in \figref{fig:instantiation}a, we show people the sample and ask them to pick words that they find as important indicators for them to judge the review sentiment. For critique-based input, as shown in \figref{fig:instantiation}b, we show the model's explanation (highlighted keywords) and ask people to critique each word's importance (agree/disagree). While the first approach can be more effortful, it can possibly obtain input for a broader set of words not limited to what is highlighted in model explanations. 

For the sampling method, we aim to select reviews where their important words show up in many other instances, which would allow good coverage of user preference information. Proposed by \citet{ribeiro2016should} as an application of LIME, SP-LIME is an example selection algorithm that selects representative instances of a data distribution. SP-LIME greedily selects examples that maximize the weight of features they contain, after omitting duplicate features.
The weight of each word is defined as the square root of the total sum of its importance across the training dataset. Using SP-LIME, we select the top 10 examples in the development set as the sample for the input step.

\(\bullet\) \textbf{How to present:} as illustrated in \figref{fig:instantiation}d, once the selective explanation is generated for an instance---which features in the raw explanations would be considered relevant or not---we \textit{gray out} irrelevant keywords. This presentation allows de-emphasizing irrelevant keywords but still maintains information about which features carried weight in the AI's prediction.

\paragraph{Belief prediction model: How to generalize from the input to generate selective explanations.} Going from the input step to the selection step requires developing computational algorithms to predict what features would be considered relevant by the user in unseen instances. To develop a model to predict such user beliefs, we use elicited user feedback on which words are relevant or not for sentiment judgment from the input stage as labels to train a word-level logistic regression model~\citep{scikit-learn}, which we refer to as the ``\textit{belief prediction model}''. In the task phase, this belief prediction model predicts whether each token in the unseen instance would be considered relevant by the user, and augment the explanation accordingly---in the instantiation, we chose to grey out tokens that are in the explanations but predicted as not relevant to sentiment judgment based on the user's belief. Since we can only gather a small amount of feedback data ($\approx 100$ labels) from each user, we intentionally choose logistic regression, which is sample-efficient due to its simplicity.
The model uses GloVe embeddings (glove-100d)~\citep{penningtonGloVeGlobalVectors2014} as features, and out-of-vocabulary words are ignored. 

With open-ended input, people would only provide positive signals (which words are relevant to sentiment judgment, both positive and negative sentiment). For critique-based input, although people would explicitly provide negative signals (which words are irrelevant to sentiment judgment), empirically we find these signals are not always reliable.
One reason is that people tend to disagree with the importance of words pointing in the opposite direction of the review sentiment (even if they are highlighted in the negative direction's color). Therefore, for both types of input, we used a strategy known as negative sampling in the literature~\citep{mikolovEfficientEstimationWord2013}, i.e., randomly sampling the same number of {\em unselected} tokens from annotation instances as negative examples, to obtain class-balanced negative signals to train the belief prediction model.

%% file: sections/study_1.tex
\section{Study 1: Selective Explanations with Self-Input}
\label{sec:study_1}

In the first experiment, we explore generating selective explanations with the instantiation described in \secref{sec:instantiation}, focusing on getting input from the \textit{individual user} with two input elicitation methods: open-ended and critique-based. The main task is to judge the sentiment of 20 movie reviews with the help of a sentiment analysis AI system, which provides its prediction for the review sentiment and explanation for its prediction. We compare participants' experience with the two paradigms to that of a control condition with the original, unselected explanations.

\input{sections/experiment}
\input{sections/results}

%% file: sections/experiment.tex
\subsection{Procedure and Participants}
\label{sec:study_design_procedure1}

\para{User study task flow.}
Participants went through four phases depending on their condition during the study: (1) consent and attention-check; 
(2) input phase (omitted for the Control condition with unselected explanations); 
(3) task phase; (4) exit survey. Participants' answers in the input phase were used to train a belief prediction model as described in Section~\ref{sec:instantiation} to generate selective explanations shown in the task phase. That is, with self-input in Study 1, each participant had a personalized belief prediction model and therefore selective explanations that varied accordingly.
Instructions to provide input and complete the movie sentiment judgment task were given before phases 2 and 3 separately. 
We added simple multiple-choice questions about the purpose of the study and what kind input they need to provide if applicable as attention-check questions. We disqualified participants who answered these questions incorrectly. 
In the exit survey, we collected basic demographic information and answers to the subjective measures described in \secref{subsection:evaluation}.
The study is approved by the IRB at the University.
Refer to the Appendix for specific details of the user study task flow.

\para{Participant information.}
Since sentiment analysis is relatively straightforward for fluent English speakers, we recruited about 40 participants for each condition from Prolific,\footnote{\url{https://www.prolific.co/}.} a popular crowdsourcing platform. To ensure high-quality responses, all participants satisfy the following three criteria: (1) residing in the United States; (2) English is their first language; (3) minimal approval rate of 95\%.
We did not allow repeated participants as the experiment follows a between-subjects design.

There were 69 male, 42 female, 5 non binary, and 2 preferred not to answer.
16 participants are aged 18-25, 55 aged 26-40, 33 aged 41-60, 12 aged over 61 and above, and 2 preferred not to answer.
Participants had diverse education background.
5 have no diploma, 19 have a diploma or an equivalent, 24 have some college credit without a degree, 8 have technical/vocational training, 59 have a Bachelor's degree or above, and 3 preferred not to answer.
Participants were paid an average wage of \$10 per hour.

\subsection{Experimental Conditions}
\label{sec:experimental_design1}

To generate selective explanations, following our framework, participants are asked to provide input based on a sample of 10 reviews (Input Phase). Participants then perform the AI-assisted decision task by judging the sentiment of 20 new movie reviews (Task Phase).

We conduct a between-subjects experiment with the following three conditions:

\begin{itemize}
    \item \textbf{Original explanations (\textit{Control}).} Participants are not asked to provide any input. In the task phase, the original explanations generated by LIME are shown together with the model prediction, as illustrated in \figref{fig:instantiation}c.
    
    \item \textbf{Selective explanations with open-ended input (\textit{Open-ended}).} In the input phase, for each review, participants are asked to write down words that they consider important indicators for their judgment of the review sentiment (\figref{fig:instantiation}a). In the task phase, participants are provided with the same AI assistance as in the \textit{Control} condition but with selective explanations instead of the original explanations.

    \item \textbf{Selective explanations with model explanation critiques (\textit{Critique-based}).} In the input phase, with the same sample as in the \textit{Open-ended} condition, participants are given the AI's explanations and asked to provide input on whether they agree that each of the highlighted keywords should be considered important for the given sentiment (see \figref{fig:instantiation}b). The task phase then shows selective explanations generated based on their critique-based input.
\end{itemize}

\paragraph{Review selection strategy.} 
The sampling strategy for the input phase (N=10) is explained in \secref{sec:framework} under ``input elicitation method''. Out of 10 reviews, 8 reviews are predicted correctly by the model, a close approximation of the model's accuracy.

For the task phase, we randomly sampled 20 movie reviews from the test set balanced for sentiment classes and model prediction correctness. We over-sampled cases where the model predictions are incorrect to better explore whether appropriate reliance happens. For Study 1, following \citet{yin2019understanding}, we opted for a fixed-seeding approach (i.e., all participants saw the same 20 reviews) to reduce variance, which turned out to be a limitation of this study, as we will discuss in the results and address in Study 2.

\subsection{Evaluation Measures}
\label{subsection:evaluation}

As discussed in \secref{sec:related_work}, informed by prior work conducting empirical studies of human-AI decision-making with explanations, we measure participants' decision accuracy (performance), reliance on AI, efficiency, and subjective perceptions about task workload, usefulness of AI assistance, and understanding of AI.

\begin{table*}
	\begin{center}
	  \begin{tabular}{ l | c c }
		\toprule
								   & {\bf Correct AI predictions} & {\bf Wrong AI predictions} \\ \hline
		{\bf Humans agree with}    & Appropriate agreement        & \textbf{Over-reliance}                  \\
		{\bf Humans disagree with} & Under-reliance                & Appropriate disagreement   \\
		\bottomrule
	  \end{tabular}
	\end{center}
	\caption{Definition of different human reliance situations based on whether the human agrees with the AI prediction and whether the AI prediction is correct. In an ideal scenario, humans will have \emph{appropriate agreement} and \emph{appropriate disagreement} with the model. Though in reality, prior work found that explanations tend to increase \emph{over-reliance}. Therefore, in this work, we focus on the measurement of \emph{over-reliance} and explore whether selective explanations can reduce it.}
	\label{tb:agreement}
  \end{table*}

\para{Accuracy.}
Human decision performance with AI assistance is measured by accuracy---percentage of reviews a participant judged correctly according to the groundtruth.

\para{Reliance.}
We are interested in investigating the effect of selective explanations on people's reliance on AI assistance, defined as the percentage of cases where people's final decision is consistent with AI prediction. Informed by prior work, we are particularly interested in whether selective explanations can reduce \textit{over-reliance}, as often found to be a pitfall of XAI for decision support. Over-reliance is defined as the percentage of cases people's decision is consistent with AI prediction when \textit{AI is incorrect}. \tabref{tb:agreement} illustrates when over-reliance happens.

\para{Efficiency.}
We measure the total elapsed time in the task phase. Elapsed time starts from the moment participants enter the evaluation phase until they complete the last review.

\para{Subjective measures.}
Our hypothesis is that selective explanations that prioritize features the recipient would consider relevant could make the explanations easier to use and more positively perceived. Meanwhile, it is also important to evaluate user experience with regard to the whole paradigm. For example, it is an open question of how providing input would impact the overall workload. We measure subjective perception with an exit survey, focusing on three categories: subjective workload, perceived usefulness of AI (with sub-measures of helpfulness, ease of task, and confidence), and understanding of AI. We list the self-rated items below, all based on a five-point Likert scale (Strongly Disagree to Strongly Agree):

\begin{itemize}[leftmargin=*,topsep=2pt]
	\item \textbf{Subjective workload}. We measure it by the average rating for three applicable items selected from NASA-TLX \citep{hart2006nasa}:
	\begin{itemize}
		\item Mental demand: I felt that the task was mentally demanding.
		\item Feelings of success (reverse item): I felt successful accomplishing what I was asked to do.
		\item Negative emotions: I was stressed, insecure, discouraged, irritated, and annoyed during the task.
	\end{itemize}
	
	\item \textbf{Perceived usefulness of AI}, with sub-measures below. We report these sub-measures separately as these items are not as established as subjective workload.
	\begin{itemize}
		\item Helpfulness: I find the information provided by the AI helpful for making movie sentiment judgments.
		\item Ease of task: Overall, the AI's assistance made the tasks easier.
		\item Confidence: If I want to make movie choices, I would feel comfortable using this AI to help me find and read positive/negative reviews.
	\end{itemize}
	
	\item \textbf{Perceived understanding of AI}, with one item: I feel I had a good understanding of how the AI makes predictions.
\end{itemize}

%% file: sections/results.tex
\subsection{Results}
\label{sec:results}

We first present results on the evaluation measures introduced in \secref{subsection:evaluation}, then dive into relevant model and user behaviors to further interpret the results. For all evaluation measures, we plot the descriptive statistics and run one-way ANOVA with the condition as the independent variable, and when significant, we conduct post-hoc Tukey's HSD test for pairwise comparisons. 

\begin{figure}[t]
\begin{subfigure}[t]{0.22\textwidth}
  \centering
  \includegraphics[trim=0 0 2.8cm 0, clip, height=4cm]{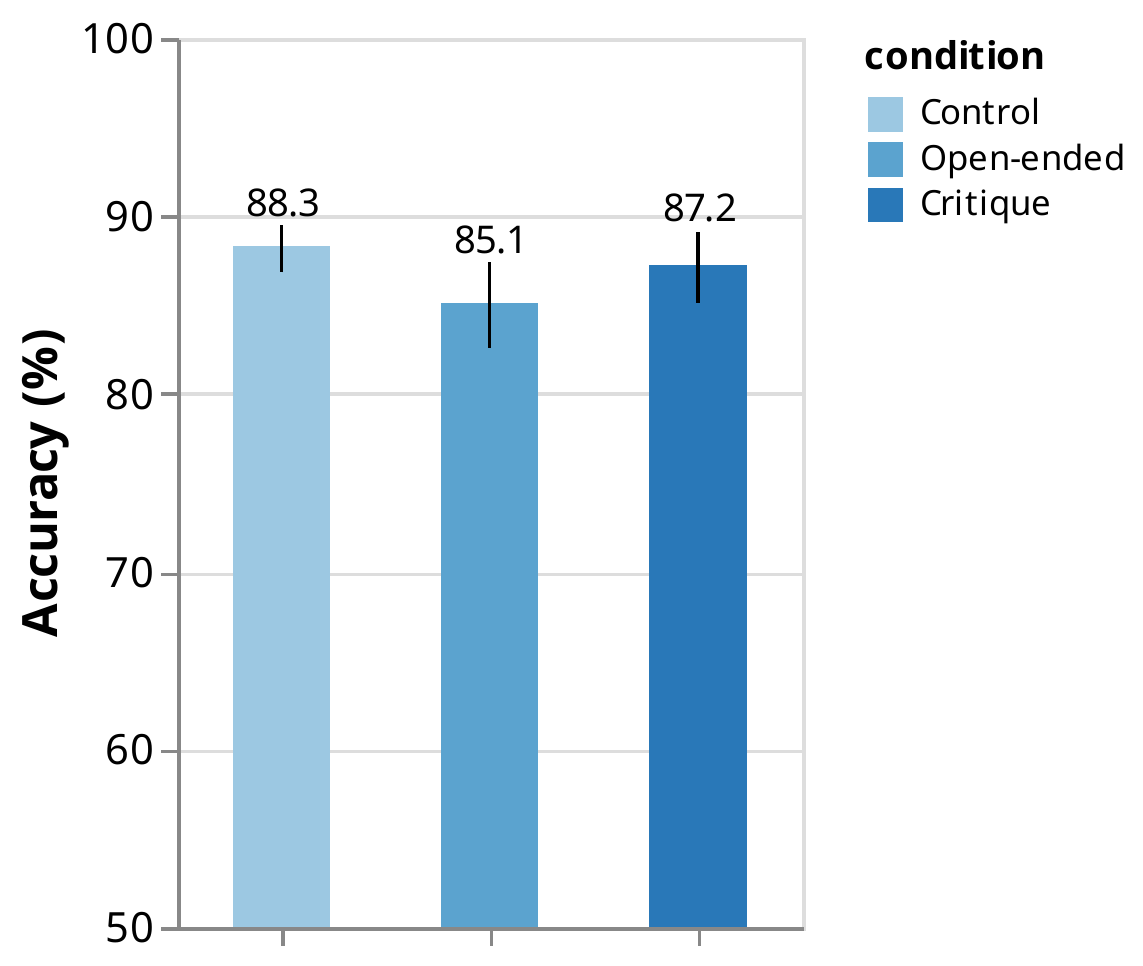}
  \caption{Performance}\label{fig:task_performance}
\end{subfigure}
\hfill
\begin{subfigure}[t]{0.36\textwidth}
  \centering
  \includegraphics[trim=0 0 2.8cm 0, clip, height=4cm]{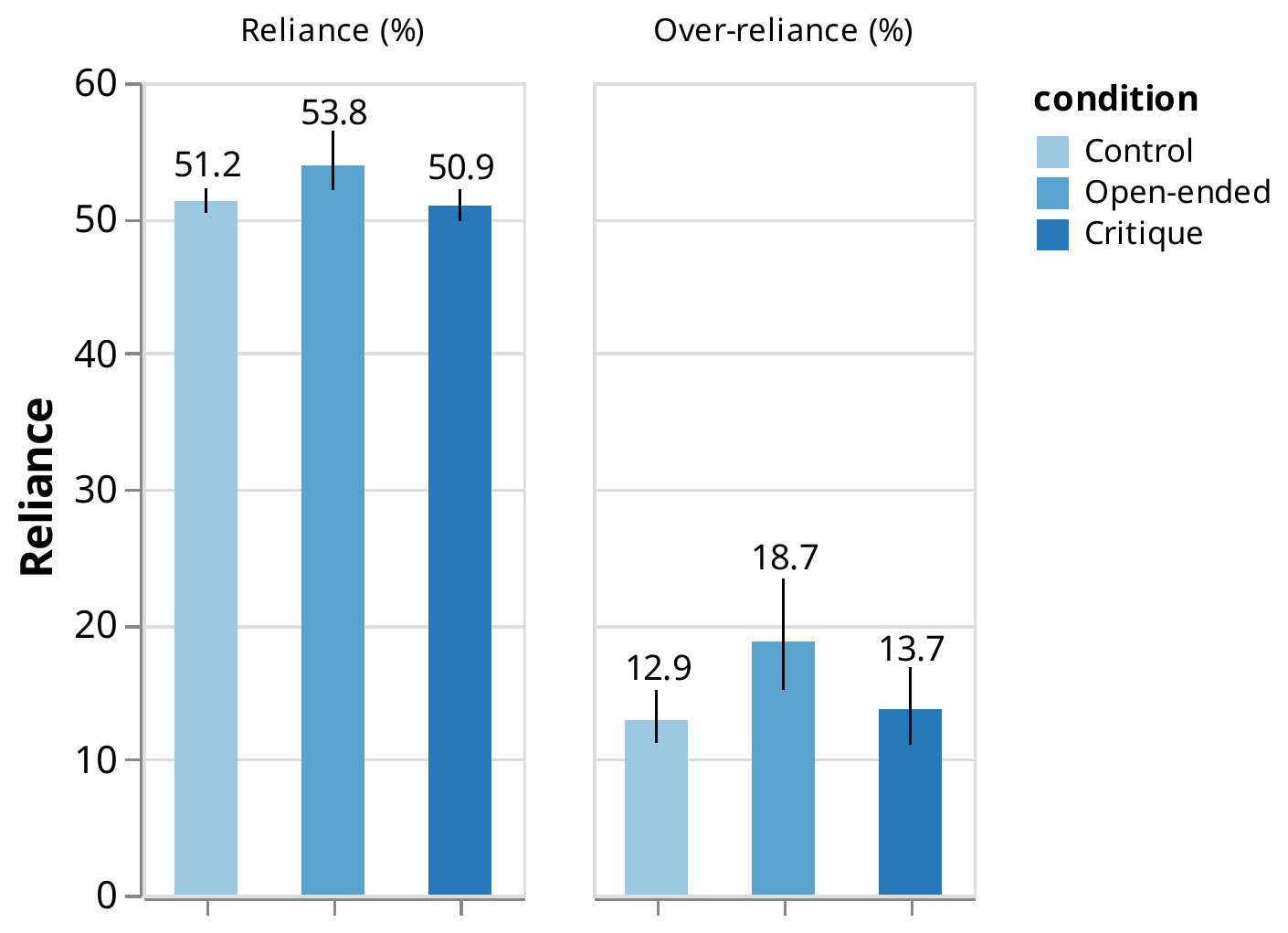}
  \caption{Reliance on model prediction}\label{fig:agreement}
\end{subfigure}
\hfill
\begin{subfigure}[t]{0.3\textwidth}
  \centering
  \includegraphics[height=4cm]{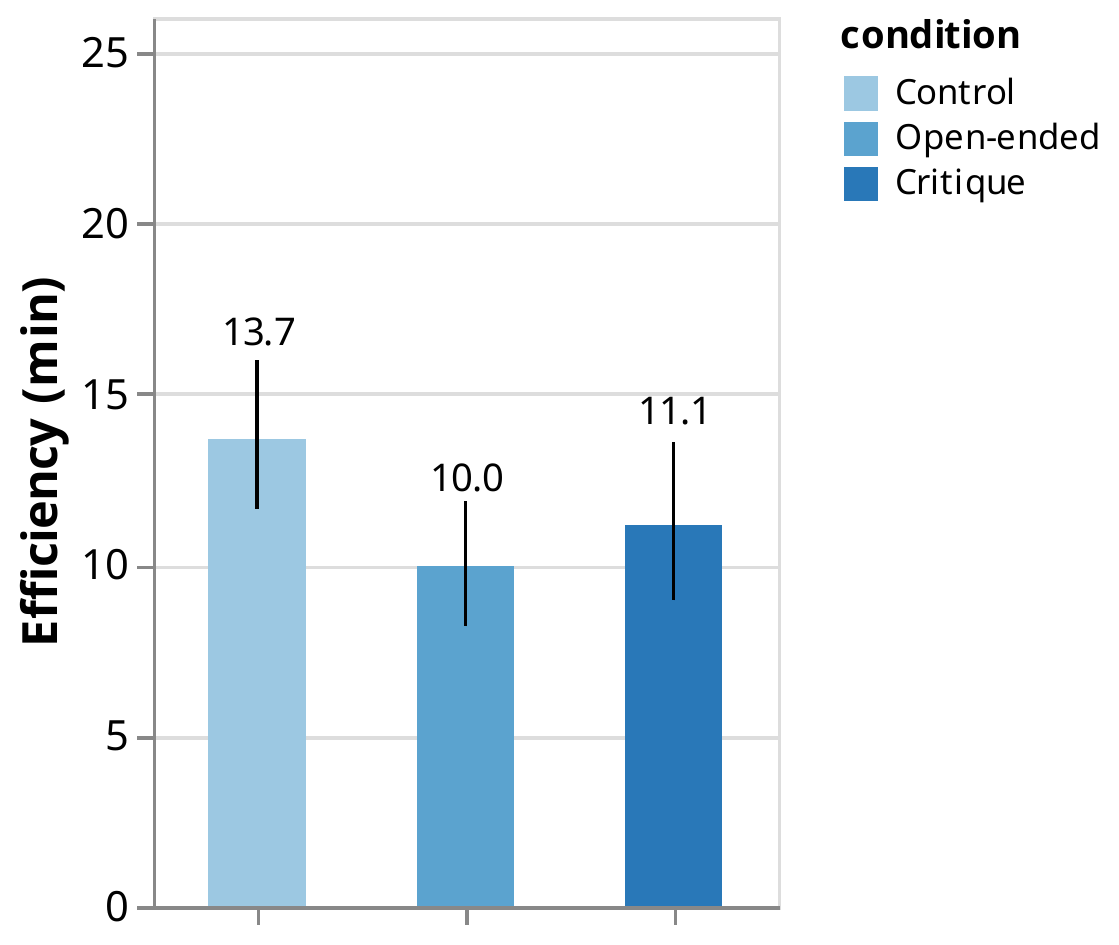}
  \caption{Time spent on the task phase}\label{fig:time}
\end{subfigure}
\\
\begin{subfigure}[t]{0.22\textwidth}
  \centering
  \includegraphics[trim=0 0 2.8cm 0, clip, height=4cm]{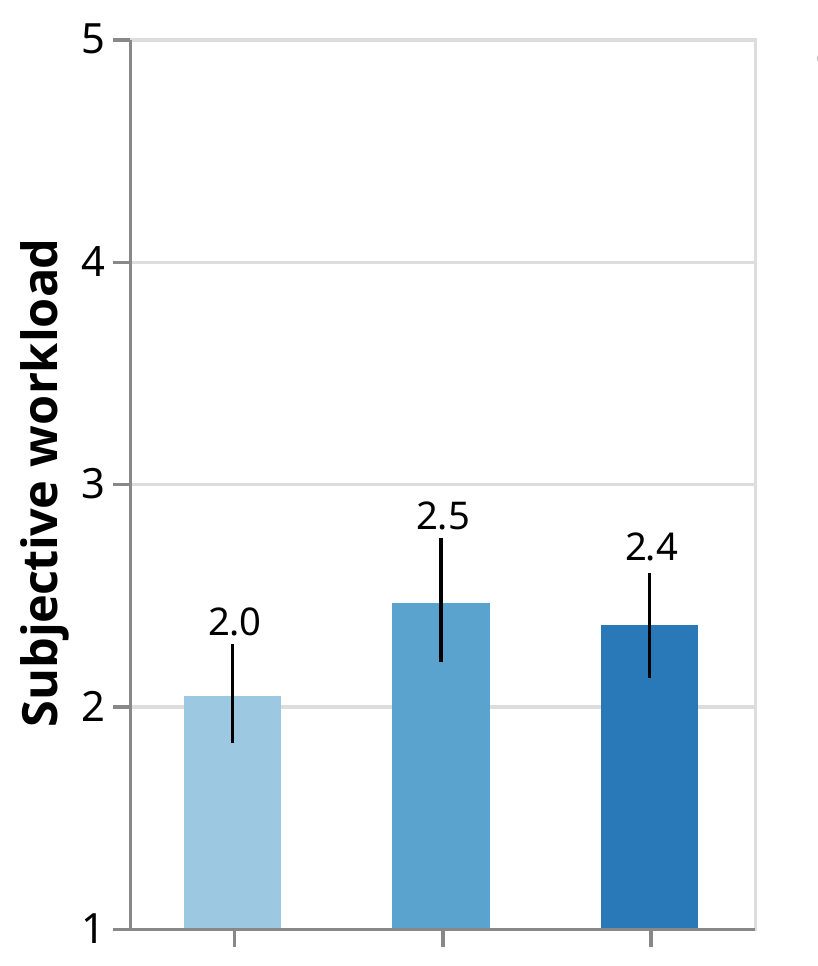}
  \caption{Subjective workload}\label{fig:subjective-workload}
\end{subfigure}
\hfill
\begin{subfigure}[t]{0.36\textwidth}
  \centering
  \includegraphics[trim=0 0 2.8cm 0, clip, height=4cm]{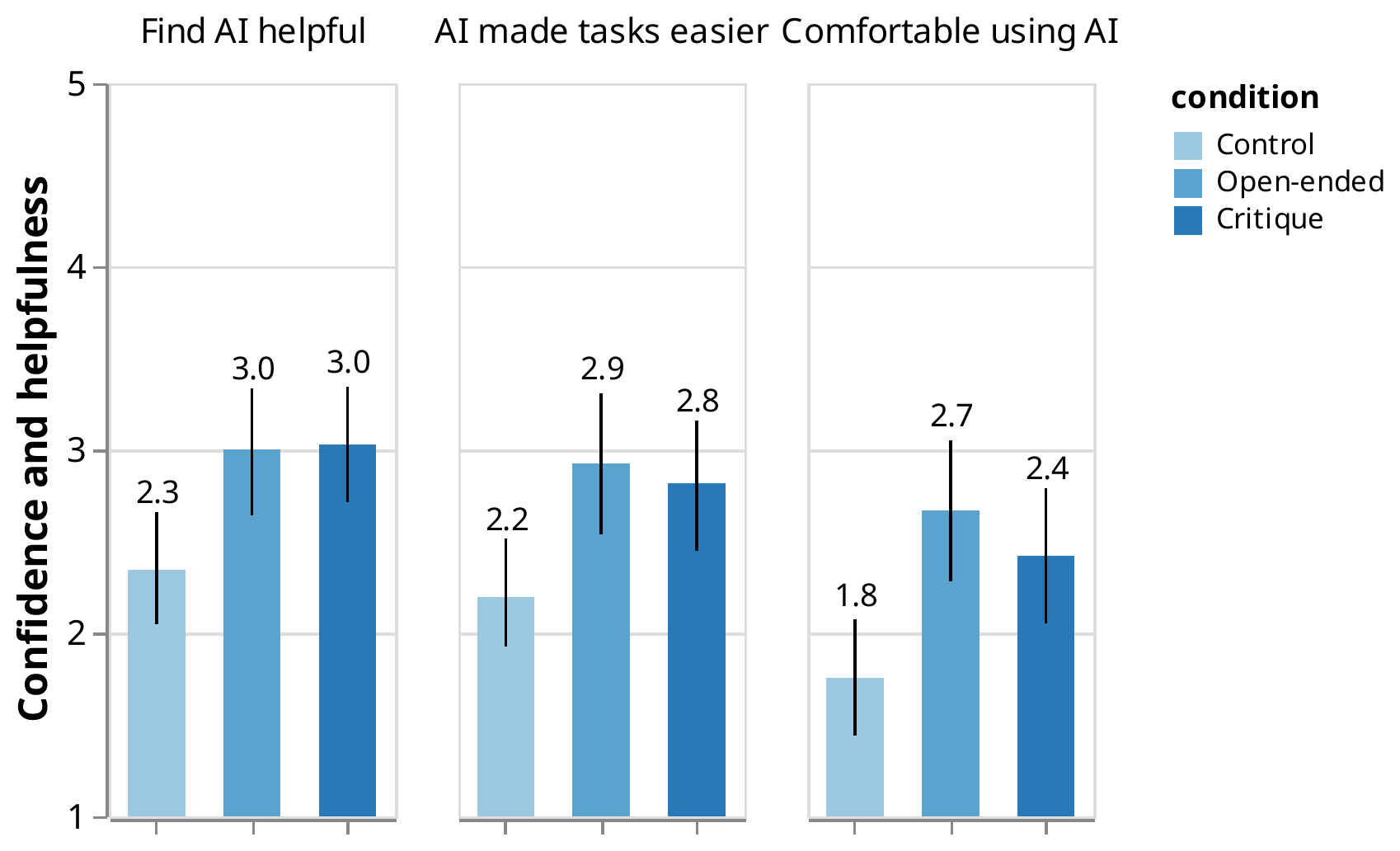}
  \caption{Confidence and helpfulness}\label{fig:confidence-and-helpfulness}
\end{subfigure}
\hfill
\begin{subfigure}[t]{0.3\textwidth}
  \includegraphics[height=4cm]{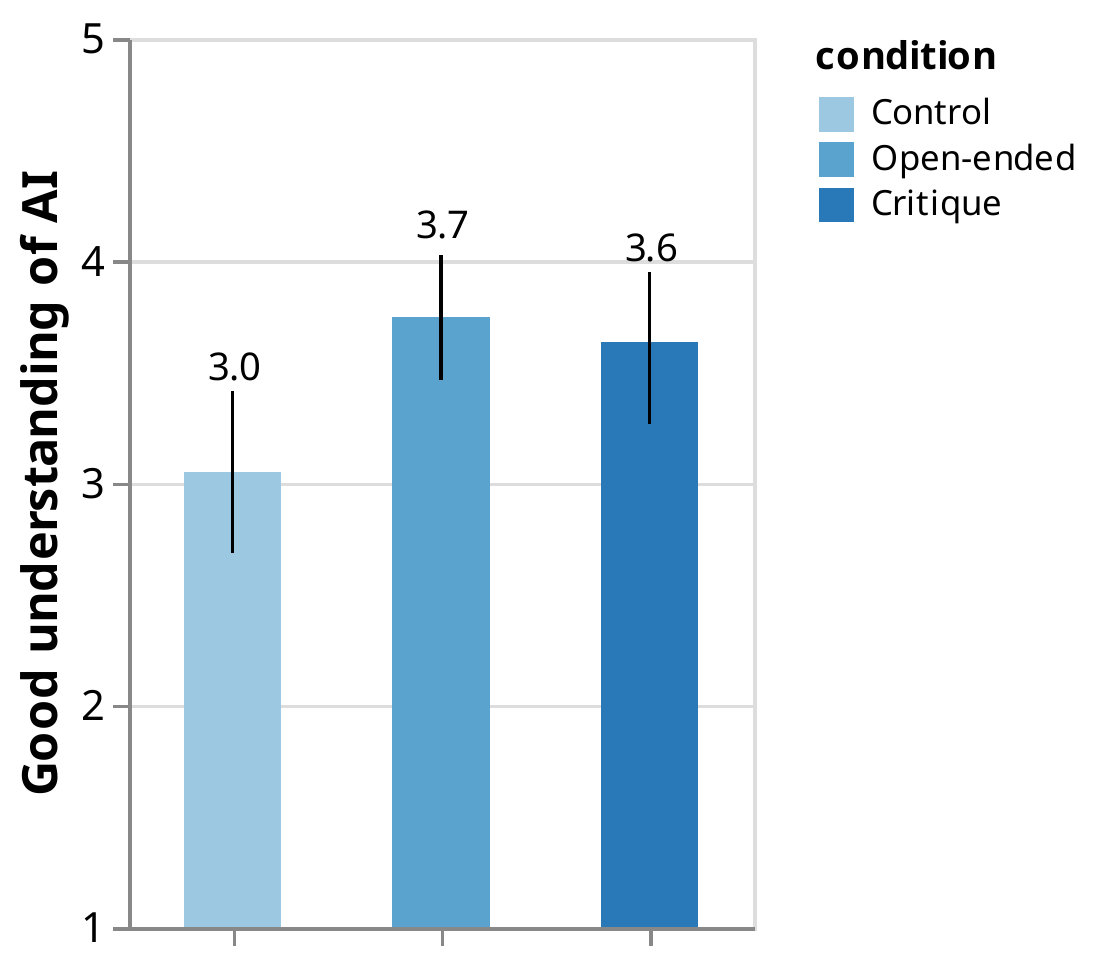}
  \caption{Perceived understanding}\label{fig:understanding}
\end{subfigure}
\caption{Results for Study 1. Error bars represent 95\% conference intervals.}
\label{fig:study1}
\end{figure}

\paragraph{Effect of selective explanations on accuracy and reliance (see \figref{fig:task_performance} and \figref{fig:agreement}).)} 
Using one-way ANOVA, we did not find a statistically significant effect of selective explanations on decision accuracy ($p = 0.09$), reliance on AI (percentage agreeing with AI, $p = 0.06$), or over-reliance on AI (percentage agreeing with AI among cases where the AI is wrong, $p = 0.06$). \figref{fig:task_performance} and \figref{fig:agreement} suggest that there is a slight, albeit non-significant, trend of the {\em Open-ended} condition resulting in the lowest accuracy and highest over-reliance. 

We note that the results of these performance-related measures should be interpreted with caution. As we will discuss later, behavior analysis showed that all participants tended to make mistakes over a small set of reviews with our fixed-seeding sampling.

\paragraph{Effect of selective explanations on efficiency (see \figref{fig:time}).}
Selective explanations led to moderately better task efficiency among participants, especially with open-ended input, as shown in \figref{fig:time}.
Although this impact is not statistically significant based on one-way ANOVA ($p = 0.06$).

\paragraph{Effect of selective explanations on subjective measures.}
\textit{Subjective workload (see \figref{fig:subjective-workload})} is measured by average ratings of the three items from NASA-TLX.  Participants who experienced the two selective explanation paradigms found the task to be more mentally demanding than those in the {\em Control} condition.
One-way ANOVA showed that selective explanations had a significant impact on mental demand ($p < 0.05$) and post-hoc Tukey's HSD showed statistical differences in both {\em Open-ended} and {\em Critique-based} conditions with the {\em Control} condition ($p < 0.05$). Looking at the sub-items, this difference is mainly due to the sub-items of being ``mentally demanding'', rather than feeling unsuccessful with the task or having negative emotions. This increased subjective workload is likely due to the extra work required in providing input.

For \textit{perceived usefulness (see \figref{fig:confidence-and-helpfulness})},
participants who were in the two conditions with selective explanations reported higher perceived usefulness of the AI tool than the {\em Control} condition across the board: higher perceived helpfulness, improved ease of the task with the AI, and higher confidence in the AI. In all three measures, one-way ANOVA revealed a significant effect ($p < 0.05$) and post-hoc Tukey's HSD suggested a statistical difference ($p < 0.05$) in both treatment conditions versus the {\em Control} condition.

Finally, for \textit{perceived understanding (see \figref{fig:understanding})}
participants who were in the two conditions with selective explanations reported a better understanding of the model than those in the {\em Control} condition. Using one-way ANOVA, selective explanations had a significant impact ($p < 0.05$), and post-hoc Tukey's HSD found the difference between {\em Control} versus {\em Critique-based} is statistically significant ($p < 0.05$), and {\em Control} versus {\em Open-ended} is marginally significant ($p = 0.05$).

In summary, while we did not find improvement in decision performance over the fixed set of reviews, we found that selective explanations with both types of self-input paradigms have a positive effect in improving perceived usefulness and understanding of the AI, a moderate effect in increasing efficiency, but also increased the overall subjective workload by requiring the additional effort of providing input.  

\subsection{Model and User Behavioral Analysis to Further Understand the Lack of Improvement on Performance}
\label{sec:more_analysis}

We conducted further analyses to unpack why we observed a slight decrease in accuracy with selective explanations in the {\em Open-ended} condition.

\begin{figure}[t]
    \begin{subfigure}[t]{0.49\textwidth}
  \centering
  \includegraphics[width=\linewidth]{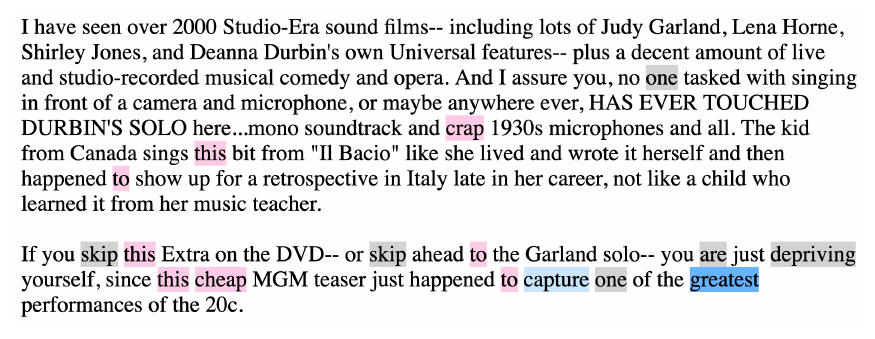}
  \caption{{\em Open-ended}}\label{fig:highlights_condition_4}
\end{subfigure} 
\begin{subfigure}[t]{0.49\textwidth}
  \centering
  \includegraphics[width=\linewidth]{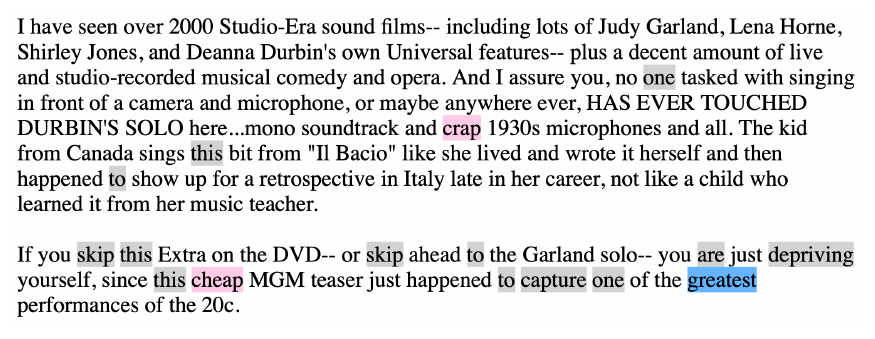}
  \caption{{\em Critique-based}}\label{fig:highlights_condition_3}
\end{subfigure} 
    \caption{Example review with selective explanations in the {\em Open-ended} condition and the {\em Critique-based} condition. For this positive review, more irrelevant words are highlighted in pink, indicating negative label, in the {\em Open-ended} conditions, while they are grayed out in the {\em Critique-based} condition.}
    \Description{Comparison between {\em Open-ended} and {\em Critique}}
    \label{fig:comparison}
\end{figure}

\paragraph{Accuracy results are biased because of two reviews.}
First, we looked at the distribution of error rates among the 20 reviews all participants saw. As shown in \figref{fig:comparison}, participants' mistakes are highly concentrated in review 0 and 19. While selective explanations based on open-ended and critique-based input reduce errors for those two reviews, they increased the errors on other reviews. The increase in errors on other reviews in open-ended input is especially prominent, which we will further address in the next paragraph. This observation suggests that our fixed-seeding sampling might have limited the generalizability of our results regarding the decision performance.

\paragraph{Input was comparatively worse in the {\em Open-ended} condition.}
We compared the quantity and quality of participants' input in the two conditions. In the {\em Critique-based} condition, participants gave relevance-positive feedback on an average of 51.9 unique words, while in {\em Open-ended} condition, the average number of unique words given was only 11.1. \tabref{tab:word-analysis} further shows the top relevant words chosen by participants in the input phase and the top misaligned words in the task phase.
Participants in the {\em Open-ended} condition overwhelmingly focus on positive-sentiment words in their feedback. In fact, none of the top 10 relevant words is relevant for negative sentiment, while the {\em Critique-based} condition identified words like ``annoying'', ''waste'', and ''worst''. This bias and lack of diversity also led to lower-quality misaligned words identified.
For instance, ``best'' and ``like'' are among the top 10 misaligned words in the {\em Open-ended} condition. 

\begin{table}[]
\centering
\small
\begin{tabular}{@{}ll|ll@{}}
\toprule
\multicolumn{2}{l|}{Top 10 selected words} & \multicolumn{2}{l}{Top 10 misaligned words} \\
Open-ended           & Critique            & Open-ended              & Critique          \\ \midrule
masterpiece          & good                & this           & this              \\
good                 & annoying            & not                     & is                \\
amazing              & fun                 & is                      & movie             \\
beautiful            & excellent           & movie                   & no                \\
happy                & best                & no                      & cast              \\
believable           & masterpiece         & best                    & and               \\
enjoyable            & waste               & like                    & story             \\
heart                & worst               & and                     & her               \\
great                & amazing             & cast                    & not               \\
real                 & beautiful           & story                   & bad               \\ \bottomrule
\end{tabular}
\caption{Top 10 selected and misaligned words in the {\em Open-ended} and {\em Critique-based} condition.}
\label{tab:word-analysis}
\end{table}

\begin{figure}[t]
\begin{subfigure}[t]{0.32\textwidth}
  \centering
  \includegraphics[width=\linewidth]{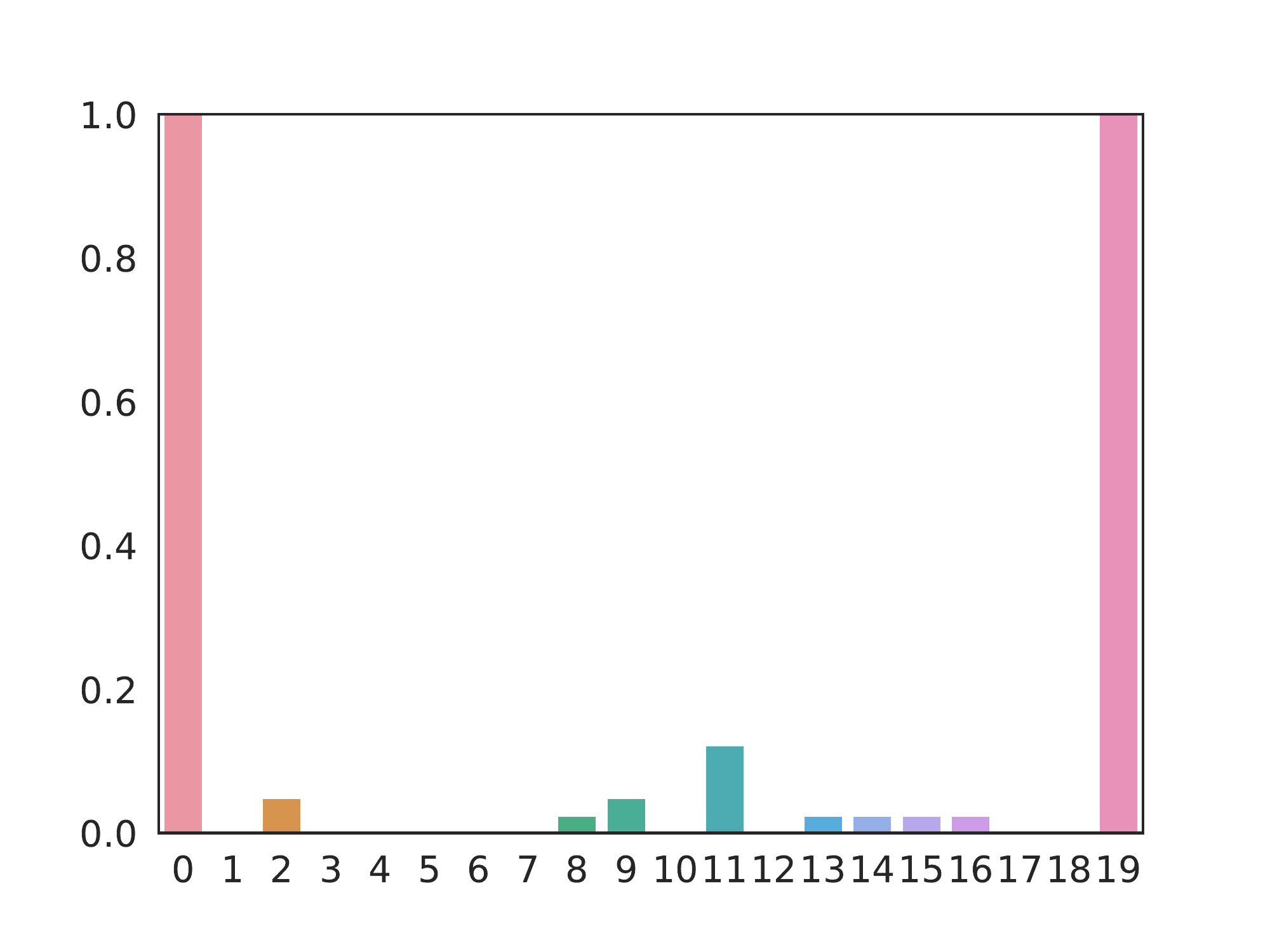}
  \caption{{\em Control}}\label{fig:error_condition_2}
\end{subfigure} 
\begin{subfigure}[t]{0.32\textwidth}
  \centering
  \includegraphics[width=\linewidth]{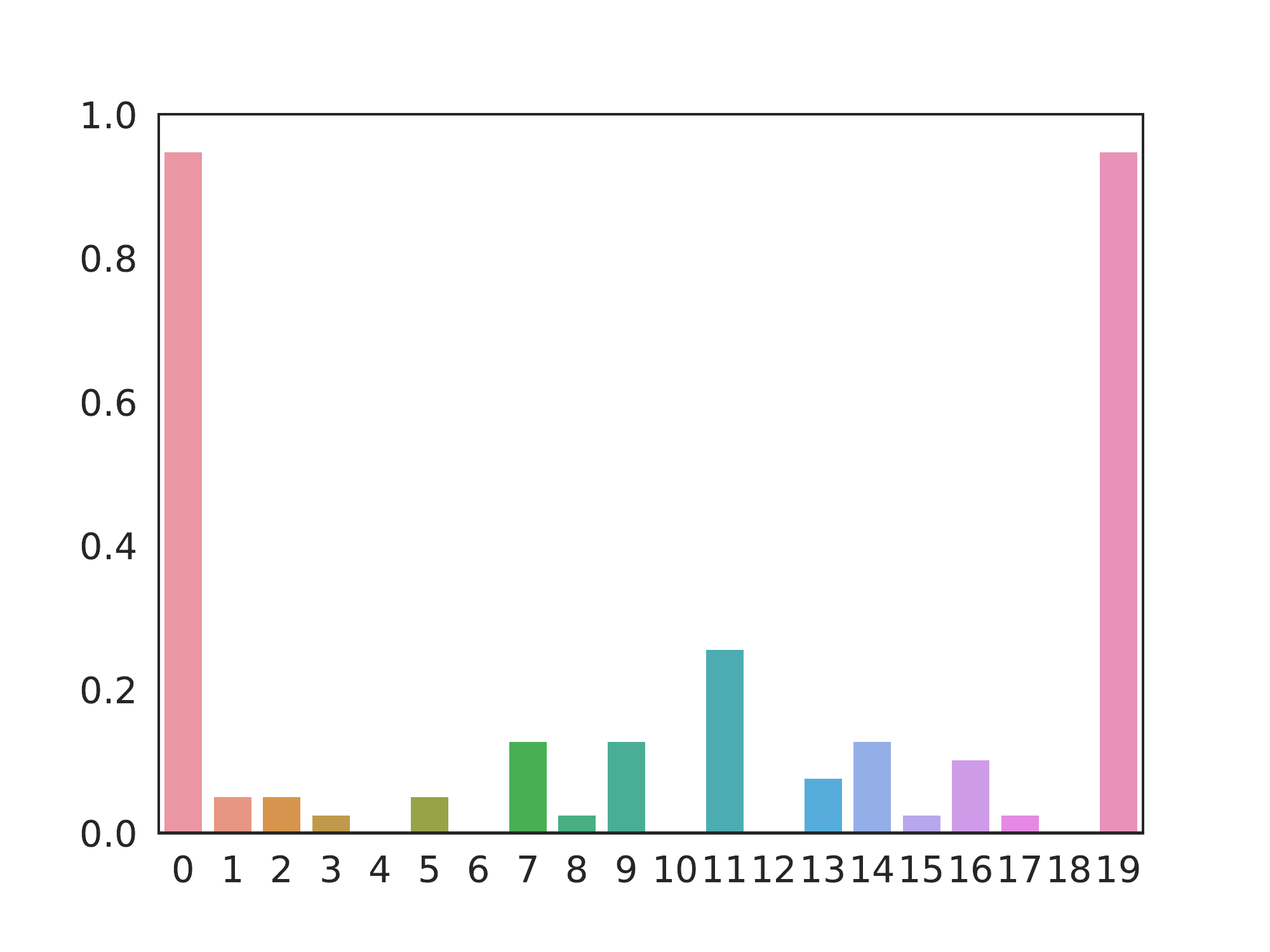}
  \caption{{\em Open-ended}}\label{fig:error_condition_4}
\end{subfigure} 
\begin{subfigure}[t]{0.32\textwidth}
  \centering
  \includegraphics[width=\linewidth]{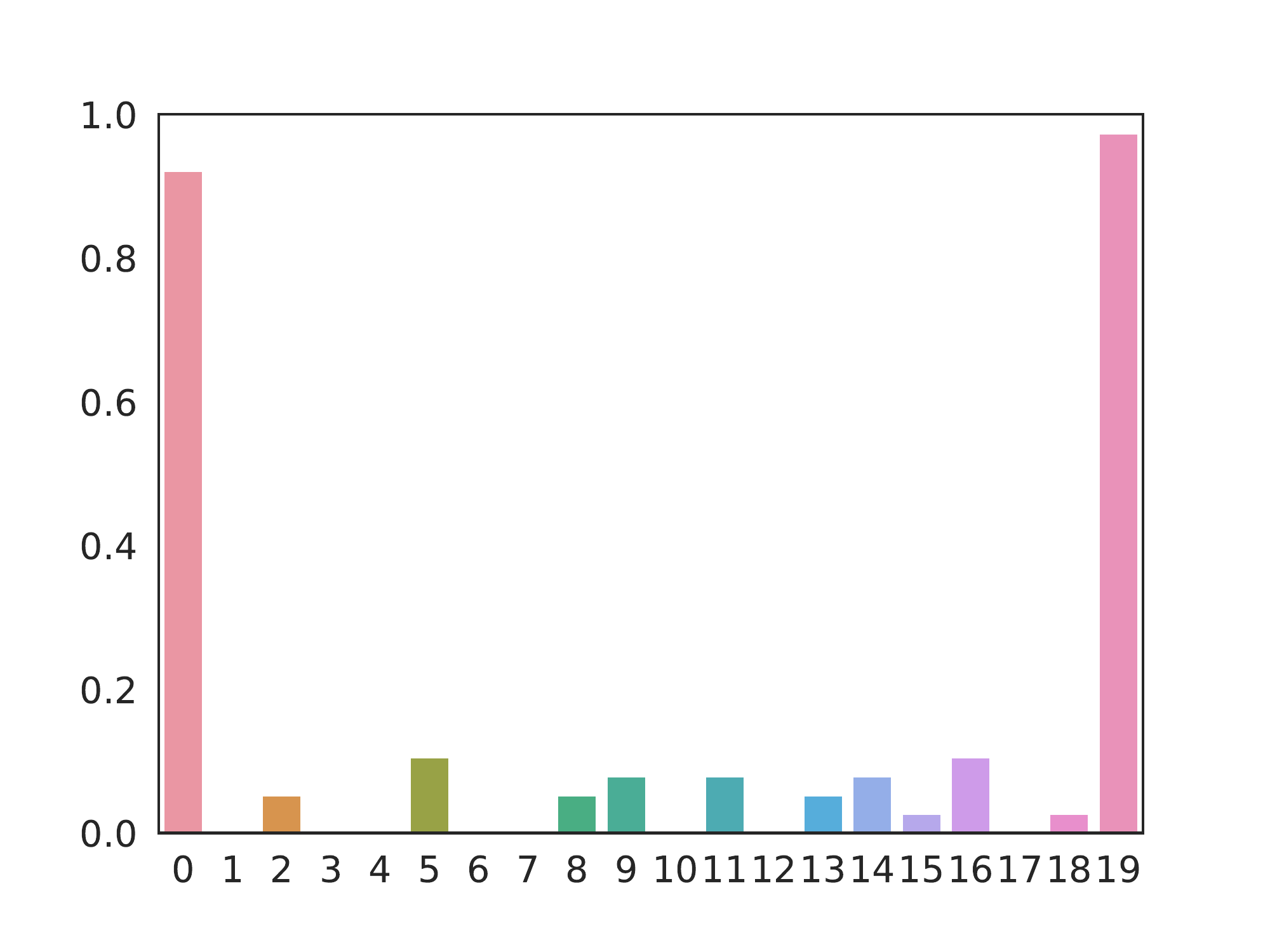}
  \caption{{\em Critique-based}}\label{fig:error_condition_3}
\end{subfigure}
\caption{Error rates grouped by review id in the three conditions.}
\label{fig:error}
\end{figure}

This lower quantity and quality of input in the open-ended may have contributed to the slight decrease in accuracy in the {\em Open-ended} condition. As shown in \figref{fig:error_condition_4} and \figref{fig:error_condition_3}, in some cases (e.g., review 11 and review 7), participants made more mistakes in the {\em Open-ended} condition. \figref{fig:comparison} shows examples for review 11 from two participants assigned to the two conditions. While the groundtruth label is positive, the AI explanation in the {\em Open-ended} condition included many irrelevant words that are highlighted as ``negative'' such as ``this'' and ``to''. In comparison, the {\em Critique-based} condition managed to gray these words out. As a result, participants were more likely to over-rely on the incorrect model prediction in the {\em Open-ended} condition and judged the review to be negative.

\paragraph{Do selective explanations increase the percentage of highlighted features supporting the groundtruth label?} We ask this question to probe on whether there is a theoretical possibility or limit for selective explanations to improve performance and decrease over-reliance. In general, if selective explanations can increase this percentage, it is more likely to nudge people toward making correct predictions consistent with the groundtruth. For example, considering when a model prediction is incorrect, the original explanation would highlight fewer features supporting the groundtruth (contradicting the prediction) than features contradicting the groundtruth (supporting the prediction). If the selective explanation can increase the percentage of the former, or even make it the majority, people may be more likely to notice keywords that support the groundtruth instead of following AI's incorrect prediction. 

\begin{figure}[t]
    \includegraphics[height=5cm]{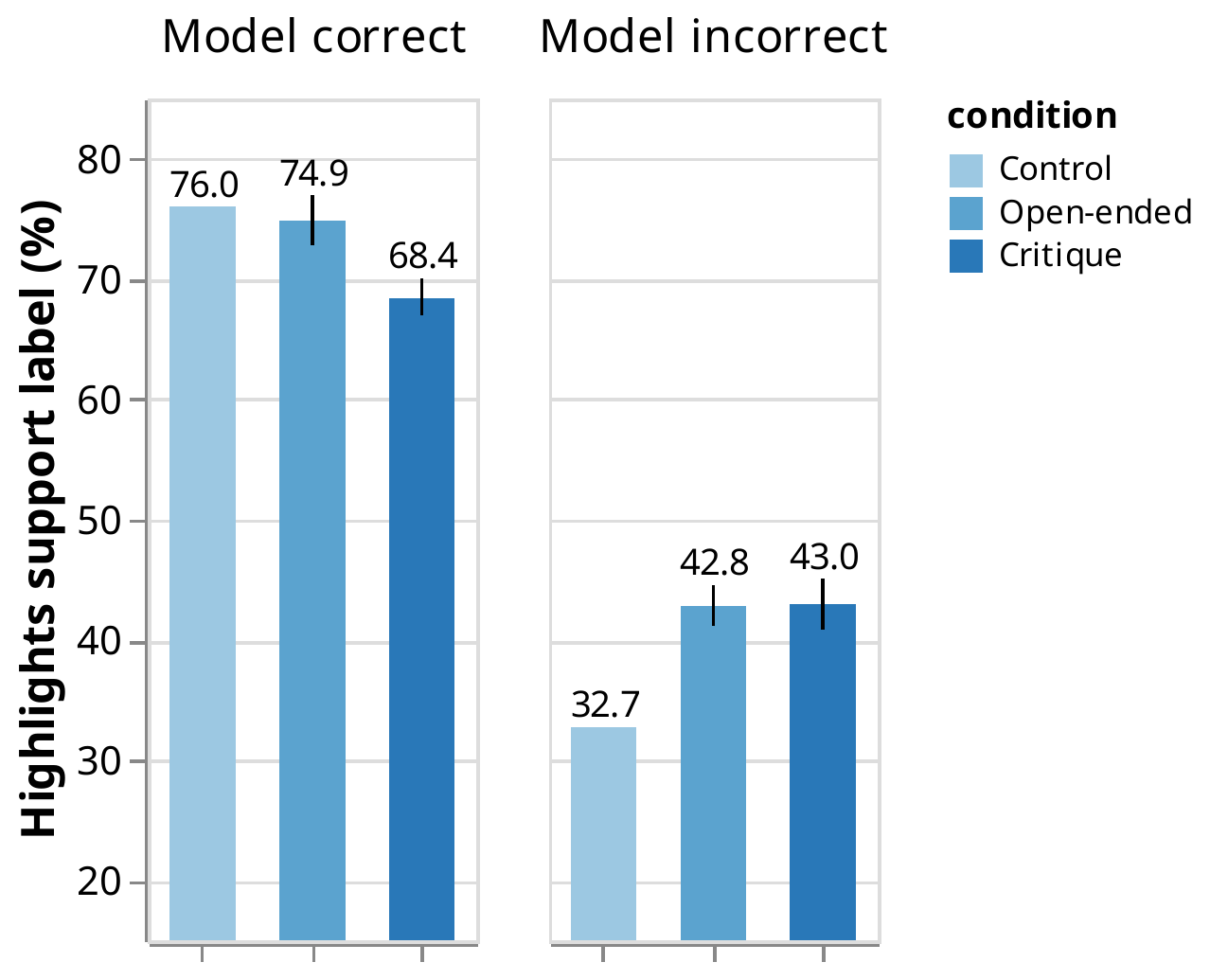}
    \caption{Percentage of highlighted features supporting the groundtruth grouped by reviews where the model made correct and incorrect predictions.}
    \label{fig:aligned-highlights1}
\end{figure}

In \figref{fig:aligned-highlights1}, we plot the percentage of highlighted words supporting the groundtruth over all highlighted words,\footnote{We count all occurrences of a unique word to account for the visual effect that each of them is highlighted.} separating reviews where the model made correct and incorrect predictions. While this percentage slightly decreased when model predictions were correct in the {\em Critique-based} condition, it consistently increased when the model predictions were incorrect with selective explanations over the control conditions. These observations suggest a theoretical possibility for selective explanation to reduce over-reliance by following the visual highlighting patterns. However, the review and keywords content may still play a greater role in people's judgment. 

In short, the additional behavioral analyses suggest that the changes to the keywords highlighting patterns through selective explanations point to a positive direction of reducing over-reliance. However, it did not translate to actual improvement, which can possibly be attributed to our sampling effect. Meanwhile, our analysis suggests that critique-based input may result in higher quantity and quality input that are more effective in identifying misaligned words than open-ended input. In Study 2, we remove the limitation of fixed sampling by using a random sampling strategy.

%% file: sections/study_2.tex
\section{Study 2: Selective Explanations with Annotator Input}
\label{sec:study_2}

In Study 2, our main goal is to explore a different input strategy by eliciting input from a panel of similar users as annotators. As discussed in Section~\ref{sec:framework}, the design choice of \textit{whose input} is an important one in practice as individuals may not afford the time and effort to provide input, as further highlighted by the increased cognitive load in Study 1. However, it is an open question whether individual users would find the input from others useful. Furthermore, in light of the limitation of fixed sampling in Study 1, we also introduce a random sampling strategy in the task phase to improve the generalizability of our results. To allow comparisons of results across the two studies and explore the robustness of Study 1 results with the fixed sample, besides an experimental condition and a control condition with random samples, we introduced another pair of them with the same fixed sample used in Study 1.

\input{sections/experiment2}

\input{sections/results2}

%% file: sections/experiment2.tex
\subsection{Study Design}
\label{sec:experimental_design2}

We conduct a between-subjects experiment with the following four conditions. To generate selective explanations with annotator input, we take all the keywords shown to participants in the {\em Critique-based} condition in Study 1 (who are ``similar'' participants recruited from the same platform with the same criteria), and take the majority vote among all  previous participants in this condition as the input data, then train the belief prediction model as described in Section~\ref{sec:instantiation}. That is, different from Study 1, where each participant had a personalized model to predict their beliefs about feature relevance based on their own input data, in Study 2, there is a fixed model for all participants. We chose to include all participants in Study 1 as the panel of annotators to avoid making arbitrary filtering decisions. The required number of annotators in practice is likely much lower. We encourage future work to explore other, more efficient, approaches to obtain annotator input.

\begin{itemize}
	\item \textbf{Random sample control (with original explanations).} This condition is similar to the control condition in Study 1, except that the reviews shown to the participants are randomly sampled while maintaining the balance of sentiment class and prediction correctness.
	\item \textbf{Random sample with selective explanations.} This condition shows selective explanations generated with annotator input but with random sampling as the condition above.
 	\item \textbf{Fixed sample control (with original explanation).} This condition is identical to the control condition in Study 1. 
	\item \textbf{Fixed sample with selective explanations.} This condition shows the same fixed sample in Study 1 and selective explanations with annotator input.
\end{itemize}

The evaluation measures and procedures are similar to Study 1, except that the input phase is removed for all conditions.

\para{Participant information.} 
Similar to Study 1, for each condition, we recruited about 40 participants from Prolific. There were 75 male, 83 female, and 3 non-binary.
31 participants are aged 18-25, 71 aged 26-40, 46 aged 41-60, 12 aged over 61 and above, and 1 preferred not to answer. In addition, participants had diverse education background: 19 are high school graduates or equivalent, 42 have some college credit without a degree, 12 have technical/vocational training, 77 have a Bachelor's degree or above, and 2 preferred not to answer. Participants were paid an average wage of \$12 per hour. Refer to \secref{sec:study_1} for details on user study task flow (the only difference lies in that there is no input phase for all conditions in Study 2).

%% file: sections/results2.tex
\subsection{Results}
\label{sec:results2}

\begin{figure}[t]
\begin{subfigure}[t]{0.23\textwidth}
  \centering
  \includegraphics[trim=0 0 2.8cm 0, clip, height=3.8cm]{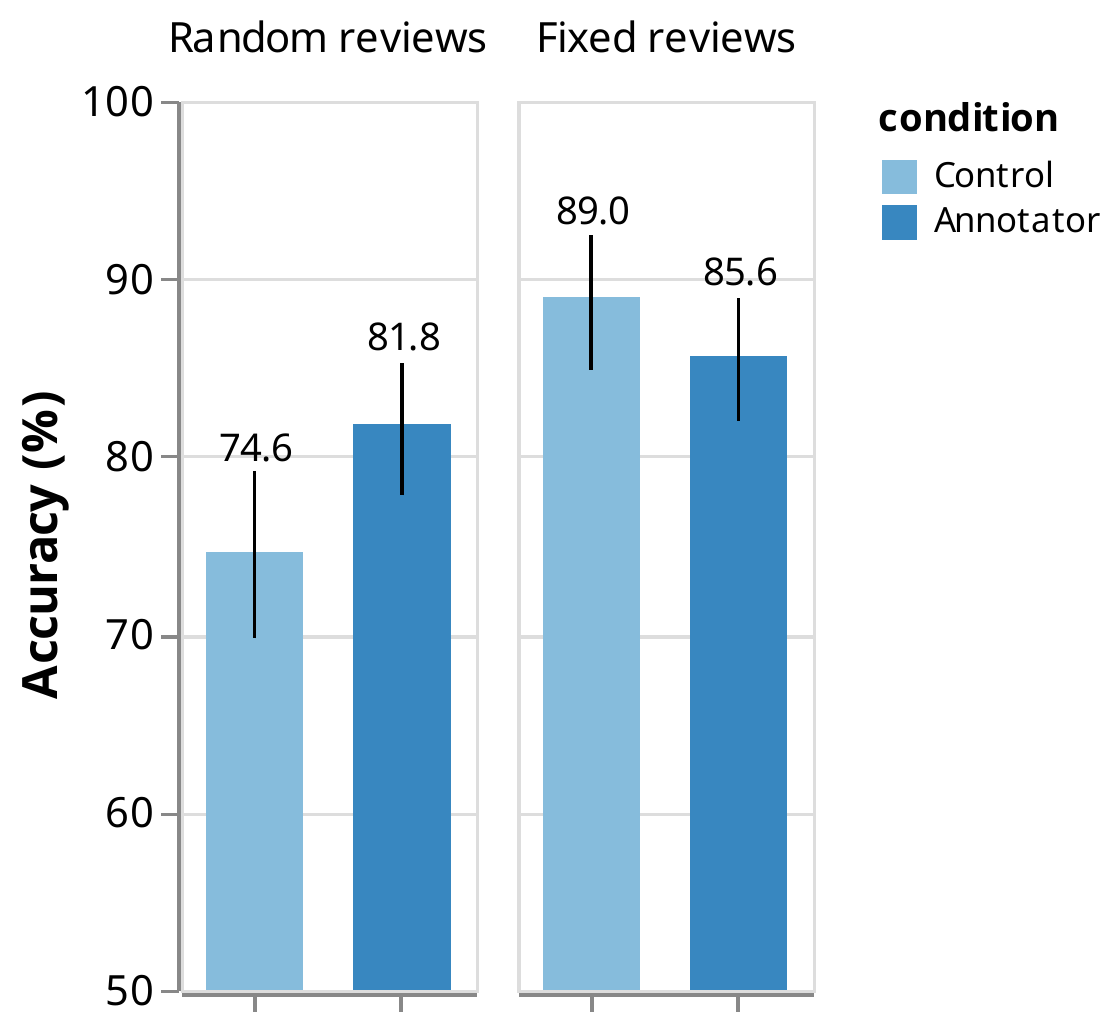}
  \caption{Performance}\label{fig:task_performance2}
\end{subfigure}
\hfill
\begin{subfigure}[t]{0.23\textwidth}
  \centering
  \includegraphics[trim=0 0 2.8cm 0, clip, height=3.8cm]{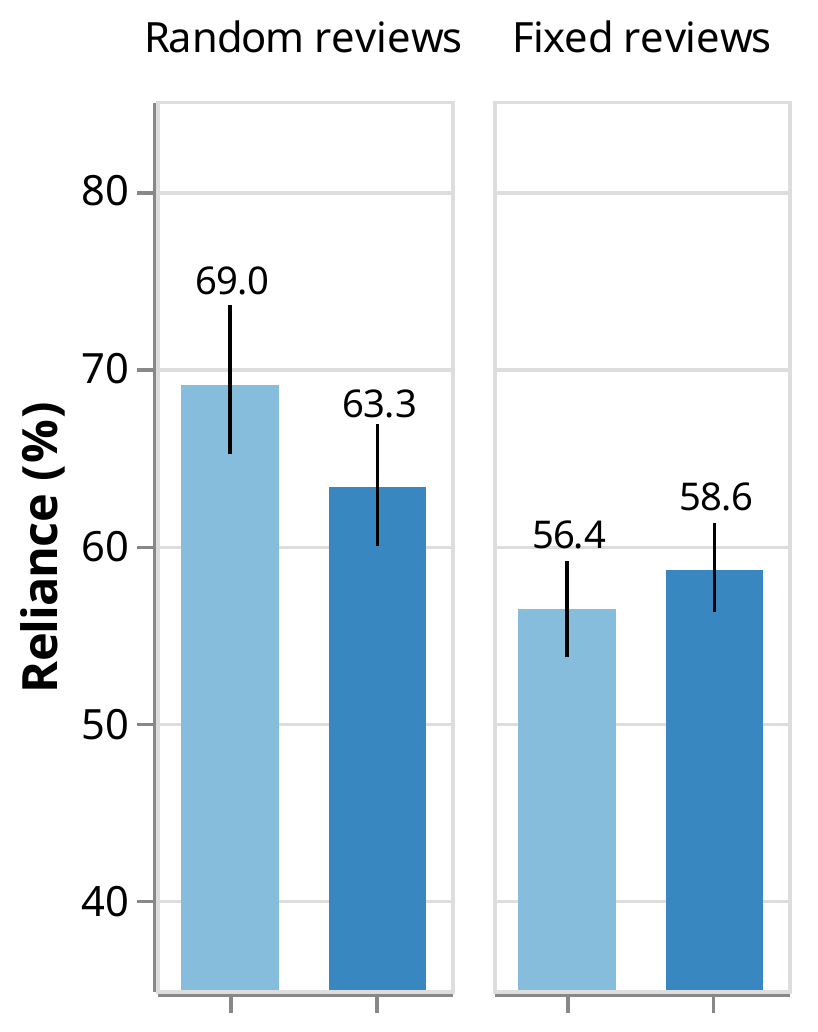}
  \caption{Reliance}\label{fig:reliance2}
\end{subfigure}
\hfill
\begin{subfigure}[t]{0.23\textwidth}%
  \centering
  \includegraphics[trim=0 0 2.8cm 0, clip, height=3.8cm]{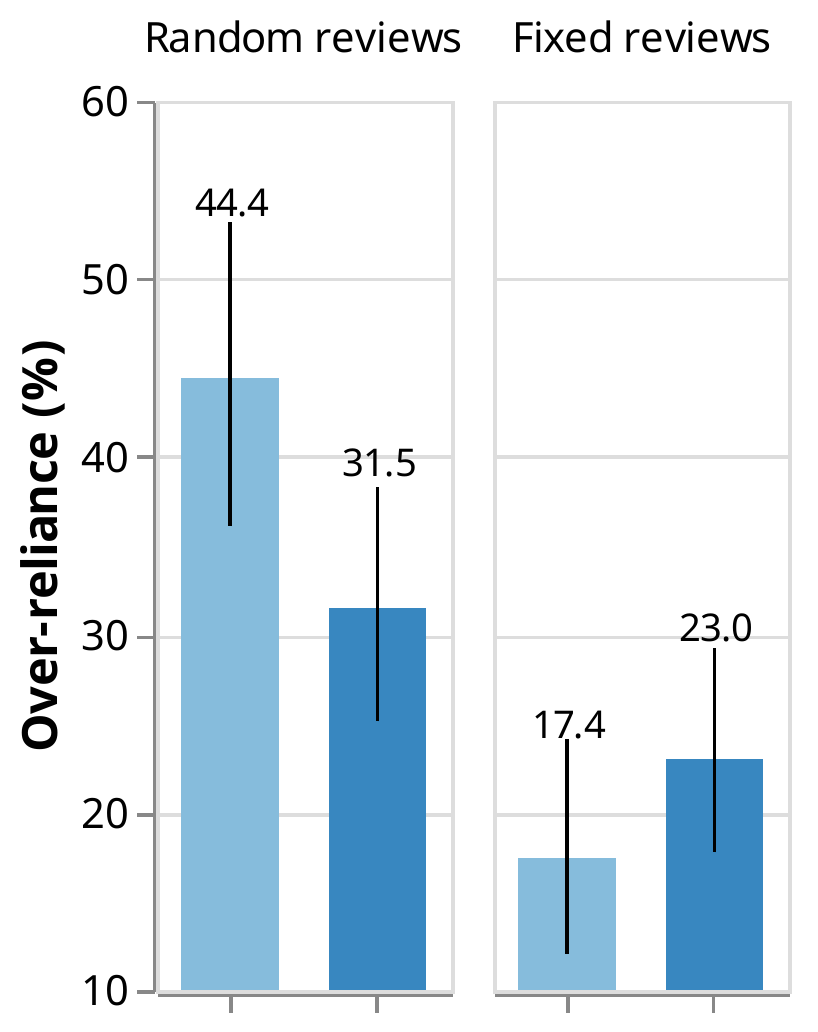}
  \caption{Over-reliance}\label{fig:over-reliance2}
\end{subfigure}
\hfill
\begin{subfigure}[t]{0.28\textwidth}
  \centering
  \includegraphics[height=3.8cm]{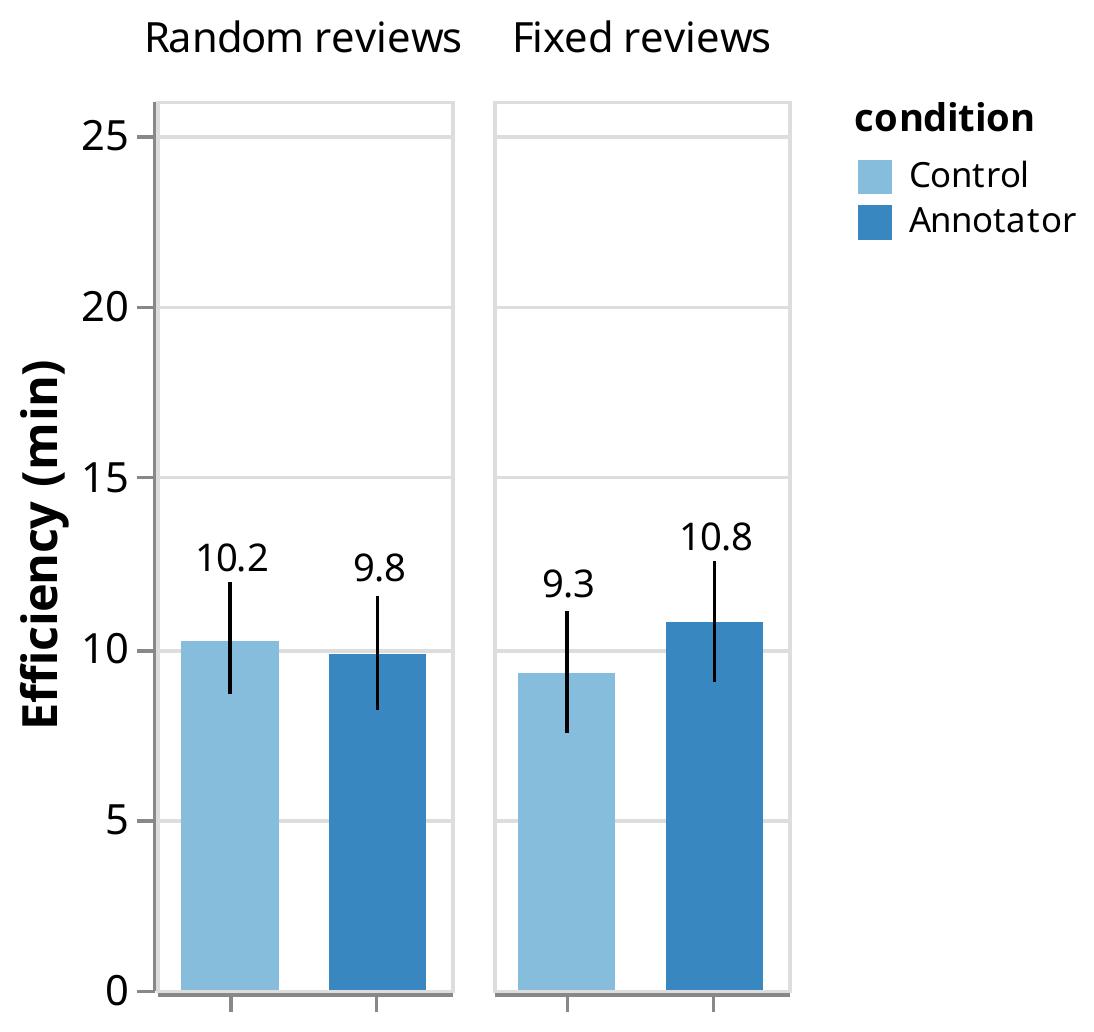}
  \caption{Time spent on the task phase}\label{fig:time2}
\end{subfigure}
\\
\begin{subfigure}[t]{0.19\textwidth}
  \centering
  \includegraphics[height=3.5cm]{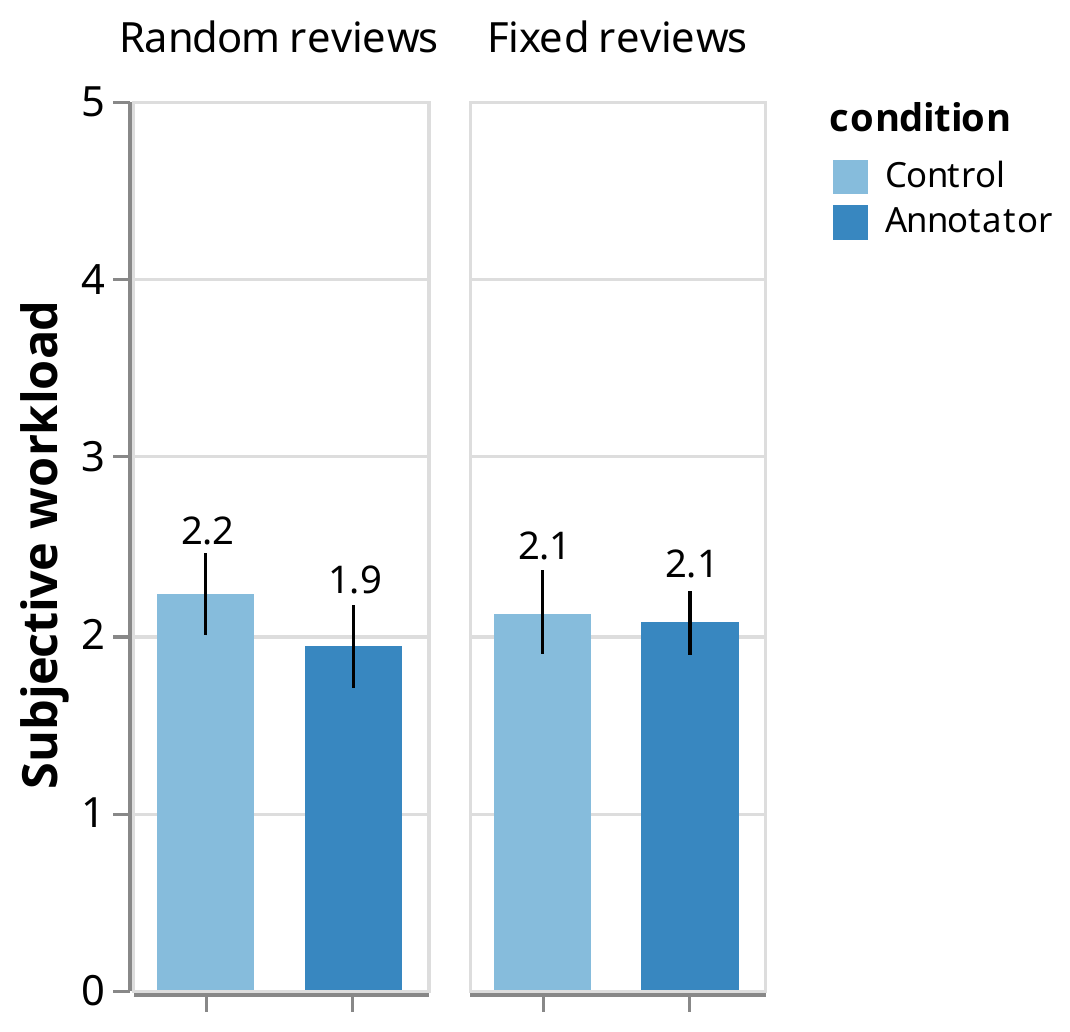}
  \caption{Subjective workload}\label{fig:subjective2}
\end{subfigure}
\hfill
\begin{subfigure}[t]{0.19\textwidth}
  \centering
  \includegraphics[height=3.5cm]{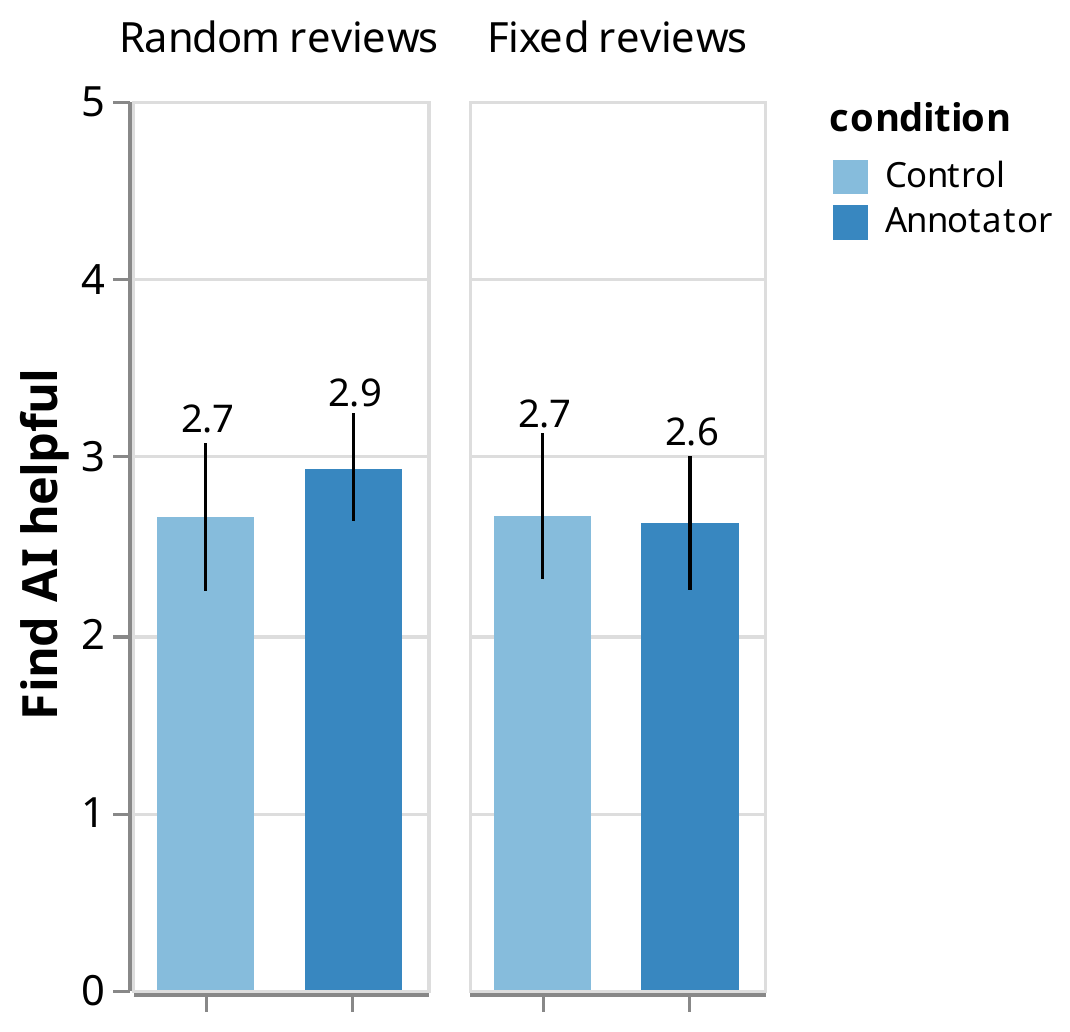}
  \caption{Find AI helpful}\label{fig:helpful2}
\end{subfigure}
\hfill
\begin{subfigure}[t]{0.19\textwidth}
  \centering
  \includegraphics[height=3.5cm]{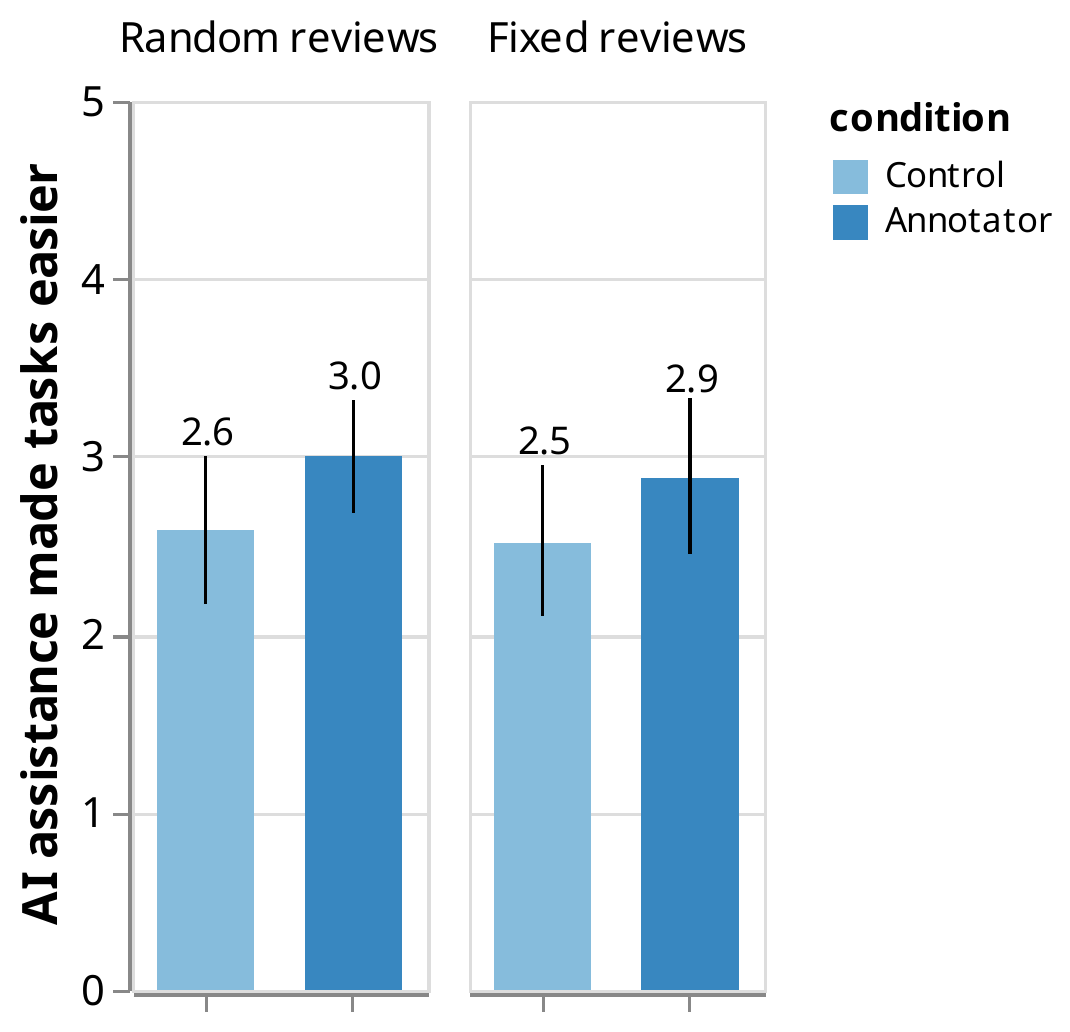}
  \caption{AI made tasks easier}\label{fig:easier2}
\end{subfigure}
\hfill
\begin{subfigure}[t]{0.19\textwidth}
  \centering
  \includegraphics[height=3.5cm]{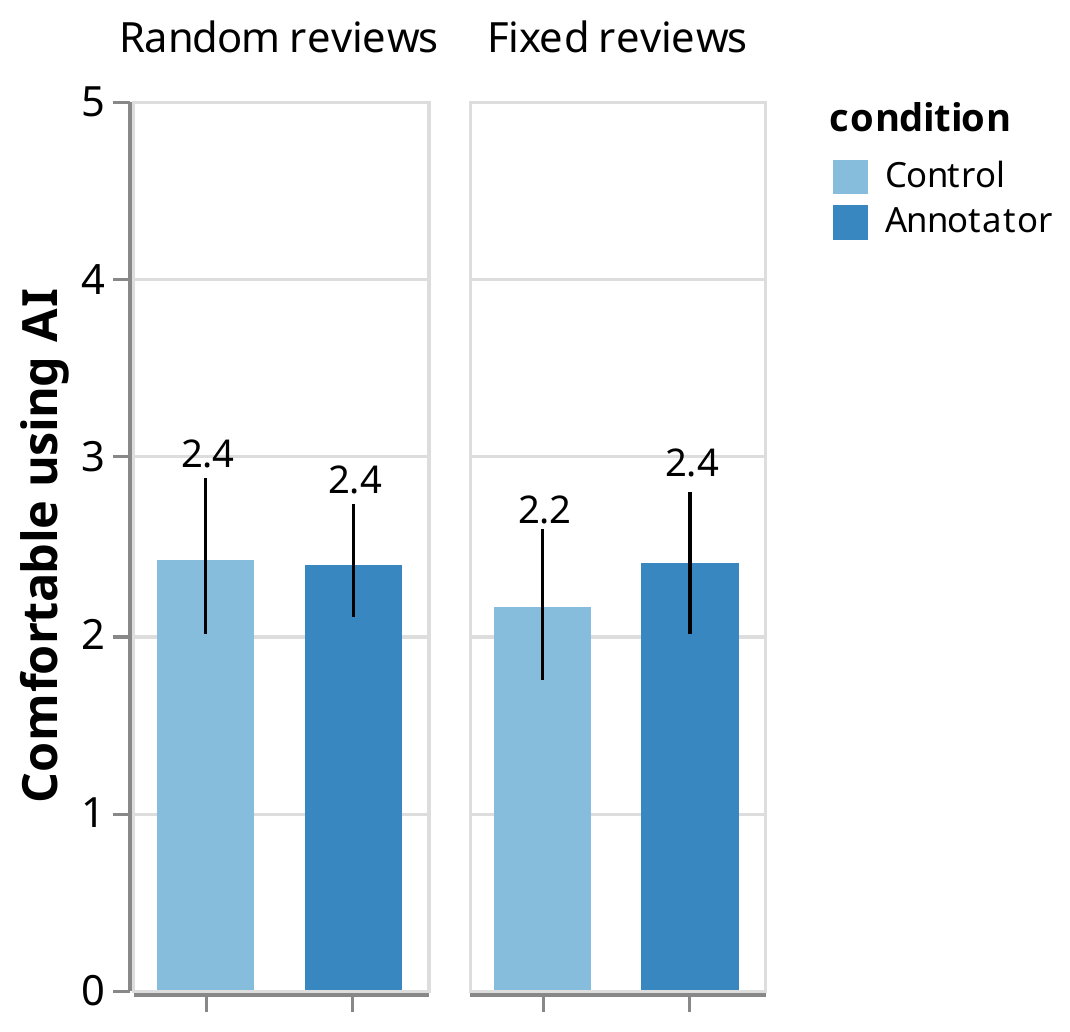}
  \caption{Comfortable using AI}\label{fig:comfortable2}
\end{subfigure}
\hfill
\begin{subfigure}[t]{0.19\textwidth}
  \centering
  \includegraphics[height=3.5cm]{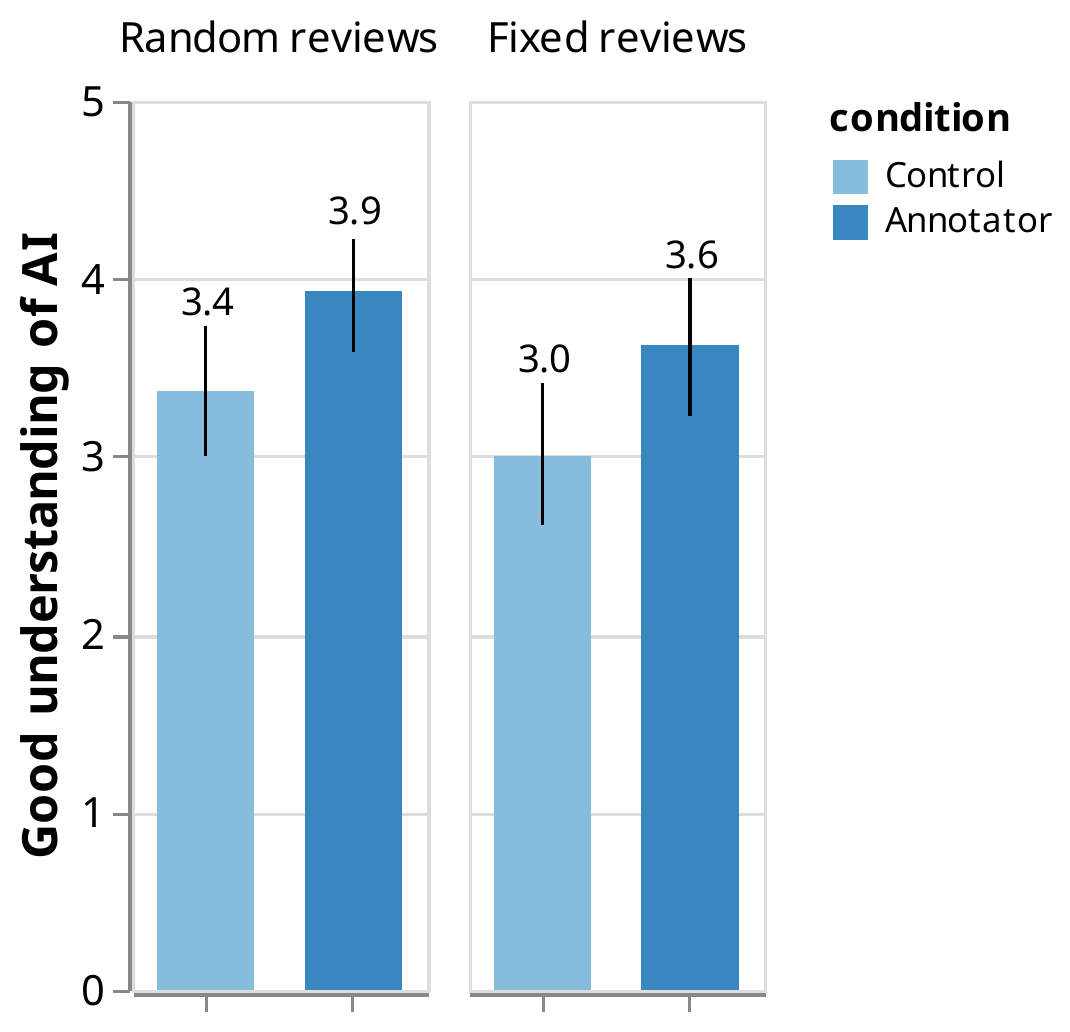}
  \caption{Perceived understanding}\label{fig:understanding2}
\end{subfigure}
\caption{Results for Study 2. Error bars represent 95\% confidence interval.
}
\label{fig:study2}
\end{figure}

We used the same evaluation measures in Study 2 as in Study 1. We start by comparing the results from the two random-sample conditions to understand the effect of selective explanations with annotator input. Then we conduct the same analyses for the two fixed-sample conditions to allow cross-study comparisons. We conduct t-tests for all measures. 

\subsubsection{Effect of Selective Explanations on Random Samples}
\label{sec:random-sample}

\paragraph{Effect of selective explanations on accuracy and reliance (see \figref{fig:task_performance2}).} 
We find a sizable improvement in accuracy from selective explanations (74.6\% vs. 81.8\%), and this difference is statistically significant ($p < 0.05$). Furthermore, we observe a substantial drop in reliance (69.0\% to 63.3\%), which can be mainly attributed to the greater drop in over-reliance from 44.4\% to 31.5\% (recall that our test samples are balanced across prediction correctness). Both differences are statistically significant ($p < 0.05$).

\begin{figure}[t]
    \centering
    \includegraphics[height=5cm]{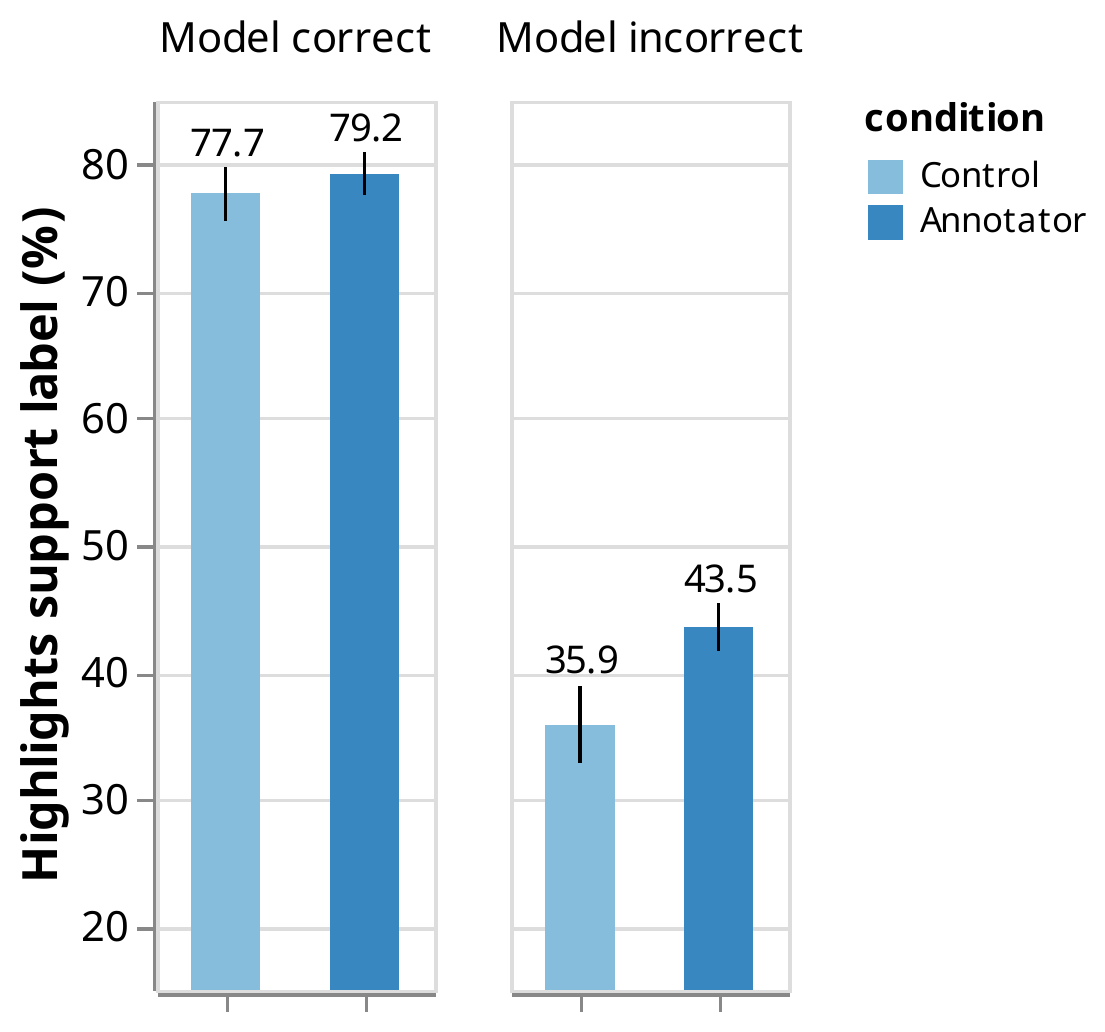}
  \caption{Percentage of highlighted words supporting the correct label in Study 2 for the random sample. The results in the fixed sample are similar to Study 1.}
  \label{fig:highlight-percentage}
\end{figure}

These results suggest that, with random sampling, selective explanations reduced over-reliance and improved overall decision performance.
Following the model behavior analysis in \secref{sec:more_analysis}, to understand the reasons for the reduced over-reliance, we examine the percentage of highlighted words that support the groundtruth label, grouped by prediction correctness. \figref{fig:highlight-percentage} shows that, with random samples, selective explanations based on annotator input substantially increases the percentage of highlighted words that support the correct label in incorrectly predicted instances from 35.9\% to 43.5\%, which could have contributed to the reduced over-reliance. Moreover, we also observe a slight increase of this percentage in correctly predicted instances, from 77.7\% to 79.2\%.

\paragraph{Effect of Selective Explanations on Efficiency (see \figref{fig:time2}).} Different from Study 1, selective explanations with annotator input virtually has no impact on efficiency ($p = 0.76$). This suggests that the moderate benefit in efficiency observed in Study 1 should be attributed to an improved familiarity with the task by completing the input phase, rather than seeing selective explanations. 

\paragraph{Effect of Selective Explanations on subjective measures (see \figref{fig:subjective2}-\ref{fig:understanding2}).} Similar to Study 1, we found that selective explanations with annotator input led to a significant improvement in the perceived understanding of AI over the control condition,  which is arguably the most important subjective measure of explanations~\cite{lai2021towards}. However, different from Study 1, we did not observe significant differences in subjective workload and perceived usefulness. This suggests that the positive effect on perceived AI usefulness observed in Study 1 should be attributed to the opportunity to provide one's own input and have control over AI outputs, rather than seeing selective explanations alone. On the other hand, removing the requirement for providing one's own input also removed the additional cognitive load observed in Study 1, suggesting a trade-off between workload and user agency to use selective explanations with others' input. 

In summary, by removing the limitation of fixed sampling in Study 1, the results in Study 2 demonstrate the promise of selective explanations in improving performance and reducing over-reliance. These results are especially exciting given the growing concerns about the XAI pitfall leading to over-reliance when the AI is wrong~\cite{bansal2021does,vasconcelos2022explanations}. We will further reflect on this result and its implications in \secref{sec:discussion}. Using selective explanations based on annotator input still consistently improved the perceived understanding of the model, and removed the additional subjective workload required by the input phase, but the positive effects on the perception of AI usefulness and efficiency are absent without the input phase to familiarize oneself with the task and have personal control over generating selective explanations. 

\subsubsection{Effect of Selective Explanations on Fixed Samples}
We now briefly discuss the results based on the two conditions with the fixed sample, mainly to compare the results with Study 1. First, similar to Study 1, we did not observe a significant difference in selective explanations on decision accuracy, reliance, and over-reliance (\figref{fig:task_performance2}), and in fact a negative trend consistent with Study 1, further confirming that the fixed sample bias results on these performance-related measures and the Study 1 results on these measures should be interpreted with caution.

Similar to the results with the random sample described above in \secref{sec:random-sample}, we found an improvement only in {\em perceived understanding of AI} ($p < 0.05$) (\figref{fig:understanding2}), but not efficiency (\figref{fig:time2}), subjective load (\figref{fig:subjective2}), or perceive AI usefulness (\figref{fig:helpful2}-\ref{fig:comfortable2}). These results suggest that these effects (and lack thereof) of selective explanations with annotator input on subjective measures are robust, and that the sampling method might have a limited impact on our results (in both studies) on subjective measures. 

%% file: sections/discussion.tex
\section{Discussion}
\label{sec:discussion}

In this work, we propose a general framework  for generating selective explanations by leveraging human input. Our framework provides a recipe for closing the gaps between AI explanation algorithms and how humans consume and provide explanations. We instantiate our framework with a text classification task and use the selective explanations in a testbed of AI-assisted decision-making. Experimental results with human subjects demonstrate the promise of selective explanations and also highlight the complexity of the design space. In this section, we further interpret the results to reflect on the underlying reasons and lessons learned, then discuss the generalizability and open questions of our framework for future work.

\subsection{Reflection on the Results}

\paragraph{Effect of selective explanations.}
We consistently find that selective explanations improve the perceived understanding of the model, which is often considered a primary goal of providing AI explanations~\cite{lai2021towards}. As shown in the example in \figref{fig:instantiation}d, by graying out irrelevant words, selective explanations are less noisy and visually sparser, concentrating on more relevant words and enabling easier sense-making of model predictions. 

We highlight the improvement in participants' decision performance and decrease in over-reliance with random sampling in Study 2. In both studies, for cases where the model is wrong, we observe an increase in the percentage of highlights supporting the groundtruth labels, thus contradicting the incorrect predictions. This suggests that, by removing irrelevant words, selective explanations are also systematically removing more ``wrongly picked'' features that contribute to the model's wrong predictions. This provides a possible path for better signaling groundtruth labels to help decision-makers avoid over-relying on the model predictions and make better decisions. This path resonates with a recent theoretical work by~\citet{chen2022machine}, which suggests that feature-based explanations can only reveal model decision boundaries (how the model makes decisions), and it is by their contrast with human intuitions about the task boundaries (which features \textit{should} contribute to the outcome) can one detect model errors. We may in fact view the gray-out words as such contrasts. 

We believe this systematic reduction of ``wrongly picked'' features and signaling of model errors should be attributed to the fact that participants in our study were able to bring in reasonable intuition about task boundaries for movie review sentiment judgment.
It is unclear whether such an outcome can be observed when the input provider knows little about the task.
Therefore, to harvest this benefit of reducing overreliance on AI, future work could consider eliciting input from \textit{domain experts} of the given decision task to generate selective explanations. However, we acknowledge that the input elicitation methods used in the current instantiation may not be optimized for generating such contrasts for wrong model predictions (e.g., the input phase saw a limited number of wrong predictions). It is also possible to create a visual design that more explicitly highlights the contrast, such as the dual-color scheme used in~\citet{boggust2022shared}, which shows human rationales in a different color. We encourage future work to explore possibilities to further enhance the effect of selective explanations on reducing over-reliance. 

\paragraph{Effect of user input.} Our results identify a few intriguing effects of user input that could have broad implications for human-in-the-loop or interactive ML work. First, we may attribute the positive effects on perceived AI usefulness and task efficiency observed in Study 1 but not Study 2 to participants providing their own input. That is, not only did participants better familiarize themselves with the task and the AI by going through an input phase, but they also felt more positively about the AI knowing that they had control over its output. While at a cost of the overall workload, these benefits of providing self-input should be broadly considered for improving user experience of AI systems. 

Second, our results suggest that when eliciting human rationales, whether to improve explanations~\cite{fenglearning,ehsan2019automated} or models~\cite{ghai2020explainable,stumpf2009interacting,cartonWhat2022}, a critique-based approach by asking for feedback to model explanations may result in better quantity and efficiency of human input over an open-ended approach. That said, it is possible to design better elicitation prompts and incentives to elicit open-ended feedback if the goal is to optimize for coverage of different features. 

\paragraph{Effect of sampling strategy.} It is worth noting that our initial results on performance suffered from the choice of sampling.
Inspired by \citet{yin2019understanding}, we chose to use a fixed sample of instances to reduce the variance. While the examples seemed representative to us, they introduced biases that limit the generalizability of our results. This observation highlights a critical challenge of studying interactions with AI-powered systems being the have high variance and uncertainty of the output space~\cite{yang2020re}. 

\subsection{Generalizability, Open Questions, and Future Directions}

\paragraph{Implications and open questions for other design choices}
Our instantiation implements only a subset of design choices for the use case of AI-supported sentiment judgment (row 1 of \tabref{tab:framework_overview}). \tabref{tab:framework_overview} lists other XAI use cases that require other selectivity goals and accordingly, modifications of the other design dimensions. Below we postulate on these design decisions and encourage future work to explore them empirically with specific XAI use cases.

To realize the abnormality goal, it is possible that providing open-ended feedback by answering ``which features should the model NOT base its decisions on'' will be especially challenging. Instead, critique-based feedback  can be elicited with a similar interface as in \figref{fig:instantiation}b but focusing on asking input about which parts of the model explanation is abnormal, either indicative of model errors (for debugging model) or surprising to the recipient (for knowledge discovery). For sampling strategy, it is possible that prioritizing cases where the model makes mistakes or different decisions from the human could be more effective for eliciting abnormality signals. For the visual presentation of selective explanations, if it is important to preserve the original explanations or the abnormal parts are relatively sparse, an alternative is to add highlights to parts that are potentially abnormal instead of greying out the rest. Lastly, depending on the use case, more sophisticated algorithms may need to be developed for generating selective explanations from human input. For example, to assist knowledge discovery, ideally the selected parts should be surprising to the user but also verifiability correct, which may pose additional requirements for the computational algorithms.

For the changeability goal, we believe the definition of ``changeability'' must be carefully operationalized according to the use case. For example, people may have different constraints on what actions they can take to improve their chance of loan approval versus their health risk. Current XAI methods, while claiming to support resource, have been criticized for false assumptions~\cite{barocas2020hidden}, including lacking mapping from changes in features to real-world actions, and negligence of interrelated changes between features or with real-life factors invisible to the model. While selective explanation offers a path for these issues by prioritizing changes that the recipient would subjectively believe to be changeable, they cannot be solved if the model features are not meaningful or lack real-world paths for change. How the changeability is operationalized should also be communicated when eliciting input, for example, by providing context such as what changes may involve and how to gauge the cost for change. Furthermore, depending on the use case, changeability could be highly personalized, and elicitation from ``a group of similar users'' may not yield useful results.

\paragraph{Extension to different models and data types.} As an augmentation approach, our framework can be applied to any existing XAI techniques that output feature-importance explanations, whether through post-hoc algorithms that generate explanations for ``black-box'' models (e.g., LIME and SHAP), or ``clear-models'' that provide feature coefficients directly (e.g., linear regression model). However, it is possible that the observed effects will be diluted if the post-hoc explanation itself is highly unfaithful to how the model actually works and introduces high noises in the explanations.

How to transfer our approach to models using other types of data than texts involves non-trivial challenges. For image data, human input could focus on giving feedback on or choosing regions of the image if high-quality image segmentation is available. For tabular data, we may need to reconsider the details of the elicitation methods. While it is possible to ask people to provide feedback for a small sample or provide their own rank in an open-ended fashion, this could become hard to manage if the feature space is large. When generalizing human input to unseen cases, the main challenge could be the non-linearity of feature importance. Therefore, we speculate that more sophisticated sampling methods may be required for tabular data to elicit human input efficiently and effectively. Furthermore, for the belief prediction, we made a simplified assumption that the role of each feature is stable across instances. A natural extension is to relax this assumption and explore the different roles of features in different instances. For text data, one possible strategy is to use contextualized word embeddings.

\paragraph{Beyond feature-importance explanations} The extension of our framework to another category of feature-based explanation---counterfactual explanations~\cite{verma2020counterfactual,wachter2017counterfactual}---is straightforward. In fact, the changeability goal is a natural fit for counterfactual explanations, which often aim to help people identify which feature they should focus on changing in order to obtain a different, often more desirable model prediction. Current counterfactual XAI techniques simply use the theoretical ``distance'' to search for features that require minimum change distance to ``flip'' the prediction to the target prediction. They can incorporate inferred recipient beliefs about changeability in the distance measure.

Future work can also explore utilizing beliefs about user preferences to augment the selection of examples in example-based explanations~\cite{guidotti2019survey,liu2021understanding,wang2021explanations}. For example, when selecting the ``most similar'' examples from the training data to explain or justify the current prediction, this similarity measure could account for which features would be considered most relevant by the recipient and assign higher weights to them. 

\paragraph{Potential issues with selective explanations}
We also encourage future work to critically examine potential issues in the assumptions underlying selective explanations. The first is a fundamental tension between selectivity and faithfulness of explanations. Whether model explanations should always be faithful is a debated topic. While it is argued that faithfulness is critical in high-stakes situations or serving actions on the model such as debugging~\cite{rudin2019stop}, some also contend that explanations do not need to be perfectly faithful (e.g., using post-hoc explanations) to provide useful information to help people make better sense of and work with the model~\cite{lipton2018mythos,liao2021human}. We further point out that the goal of selectivity is not to deceive but to help people better process information without being overwhelmed, and people should maintain control of what they wish to see. Under our framework (Section~\ref{sec:instantiation}), we also recommended multiple visual presentation choices that still preserve the original content with less tradeoff of faithfulness, such as adding highlights to original explanations. That being said, future work should examine in what situations selective explanation can result in missing information and what are the risks.

A second set of issues are related to who can provide input and who will be disadvantaged. As Study 1 shows, providing one's own input adds cognitive load, and in practice, not every individual can afford the time and resources to do so. Even if opportunities are given to everyone, the quality of input may vary by their expertise, time available, and other individual factors, which can result in inequality of benefits they can harness from selective explanations. Future work should explore how to narrow the gaps in input quality through more efficient and better-designed elicitation methods. Our Study 2 provides positive evidence for eliciting input from a panel of annotators, which can eliminate the burden for individual users and possibly mitigate the inequality issue. However, open questions remain on how to choose such a panel to be representative and inclusive, what are the risks of misrepresentation, and how to regulate system developers to avoid intentional misrepresentation.

%% file: sections/appendix.tex
\section{Appendix}
\label{sec:appendix}

\subsection{User Study Task Flow}

Generally, the participants went through four phases during the study and we will describe each phase in detail. Refer to the figures for details on the interface.

\begin{figure*}[t]
    \centering
    \includegraphics[angle=-90,width=0.95\textwidth]{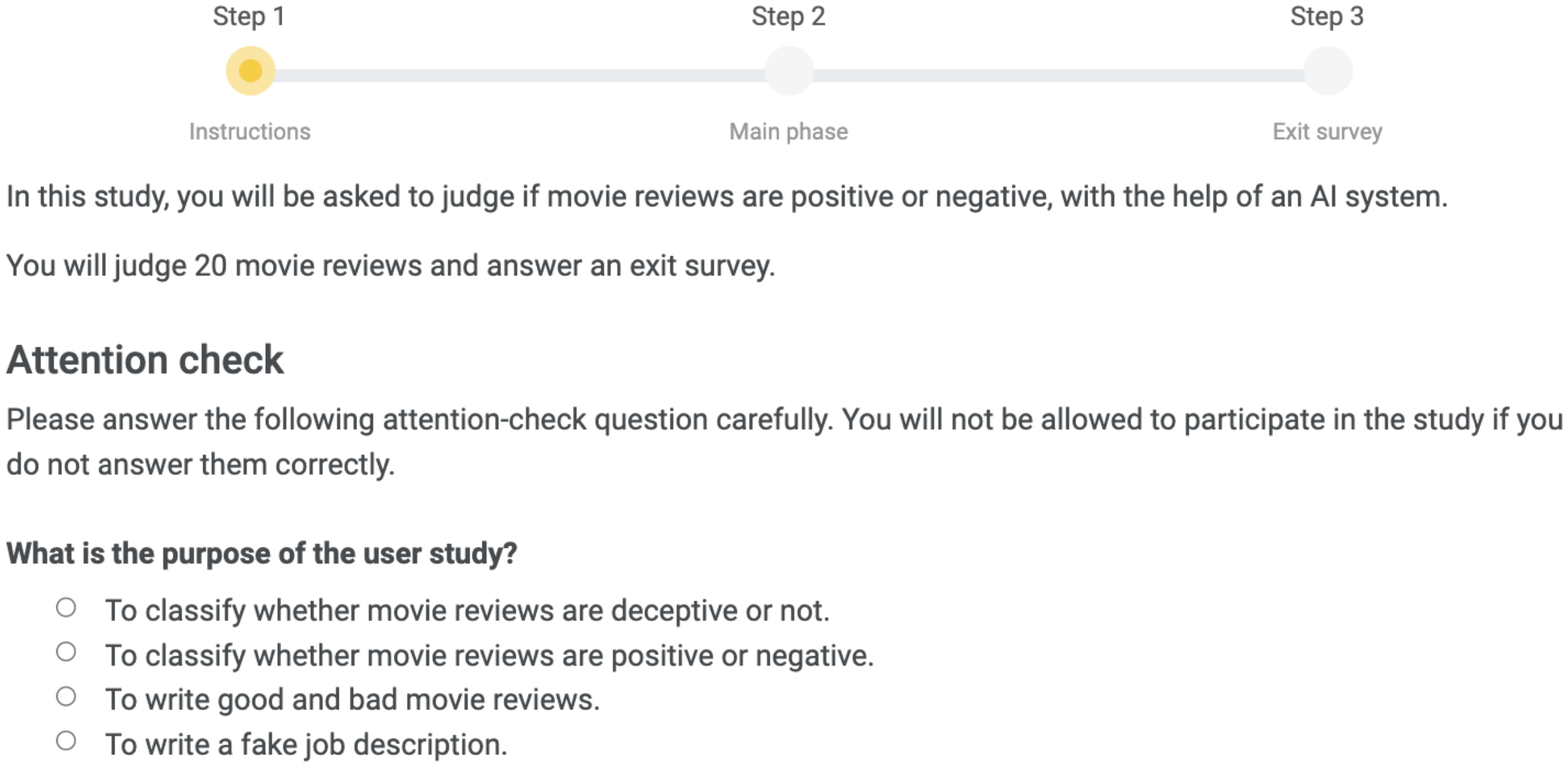}
    \caption{The participant is briefed on what is expected of the task. This figure shows the instruction of a participant whose task is to perform the prediction task only.}
    \label{fig:user_study_flow_attention1}
\end{figure*}

\begin{figure*}[t]
    \centering
    \includegraphics[width=0.95\textwidth]{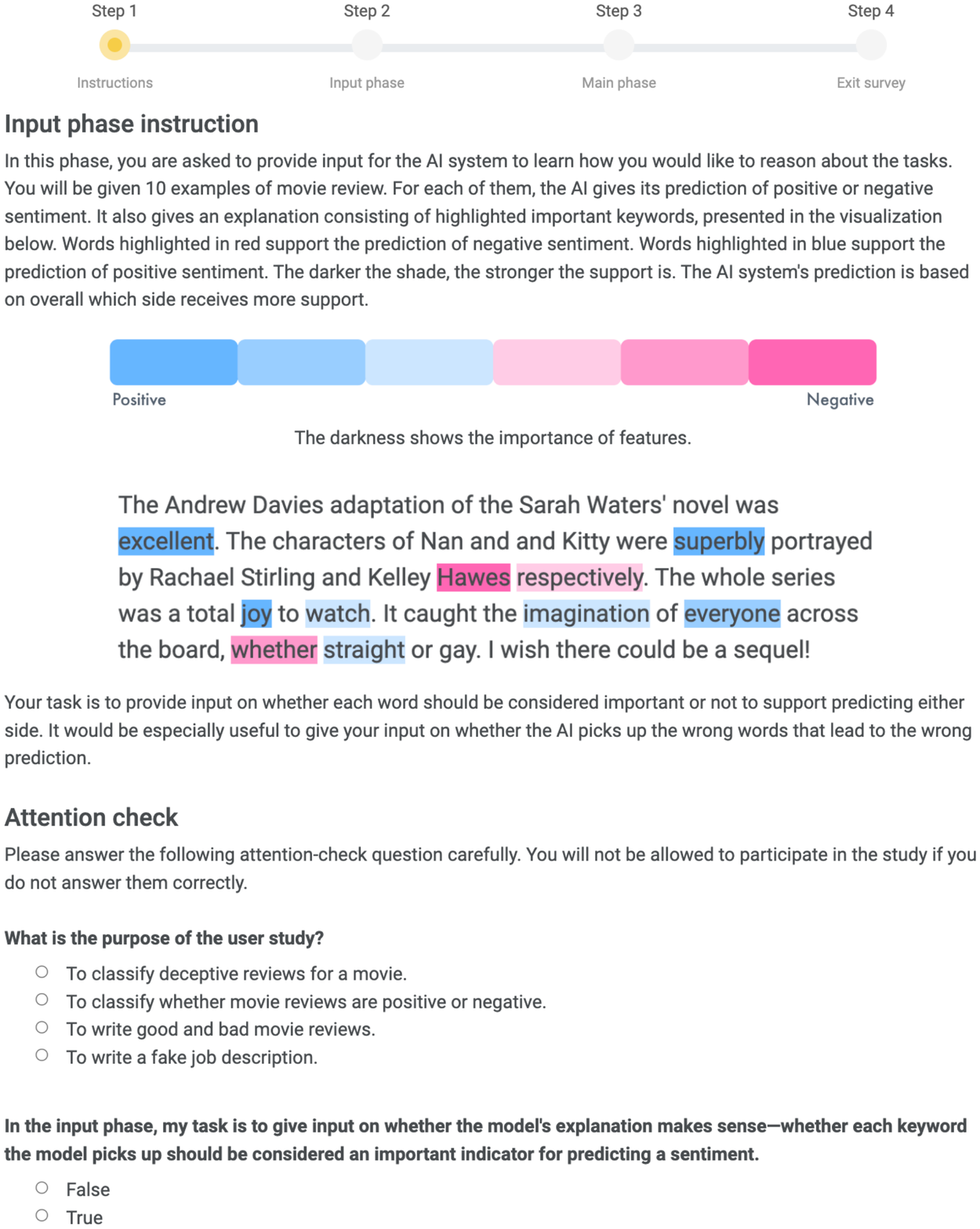}
    \caption{The participant is briefed on what is expected of the task. This figure shows the instruction of a participant who is tasked to provide input on their beliefs in the explanations and perform the prediction task.}
    \label{fig:user_study_flow_attention2}
\end{figure*}

\begin{figure*}[t]
    \centering
    \includegraphics[angle=-90,width=0.95\textwidth]{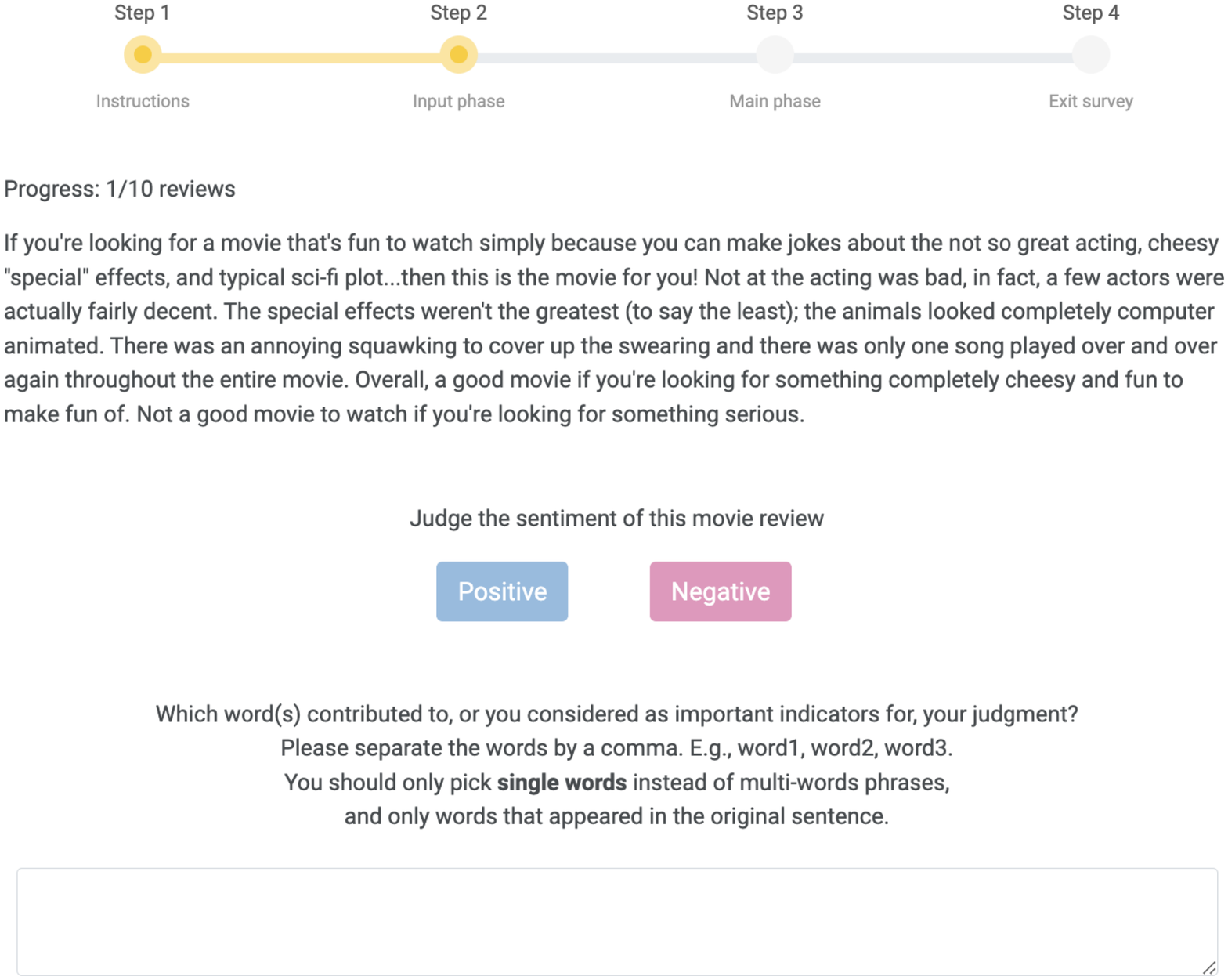}
    \caption{There are two ways of providing input. This figure shows the interface of selective explanations with open-ended input (Open-ended).}
    \label{fig:user_study_flow_input_open_ended}
\end{figure*}

\begin{figure*}[t]
    \centering
    \includegraphics[width=0.95\textwidth]{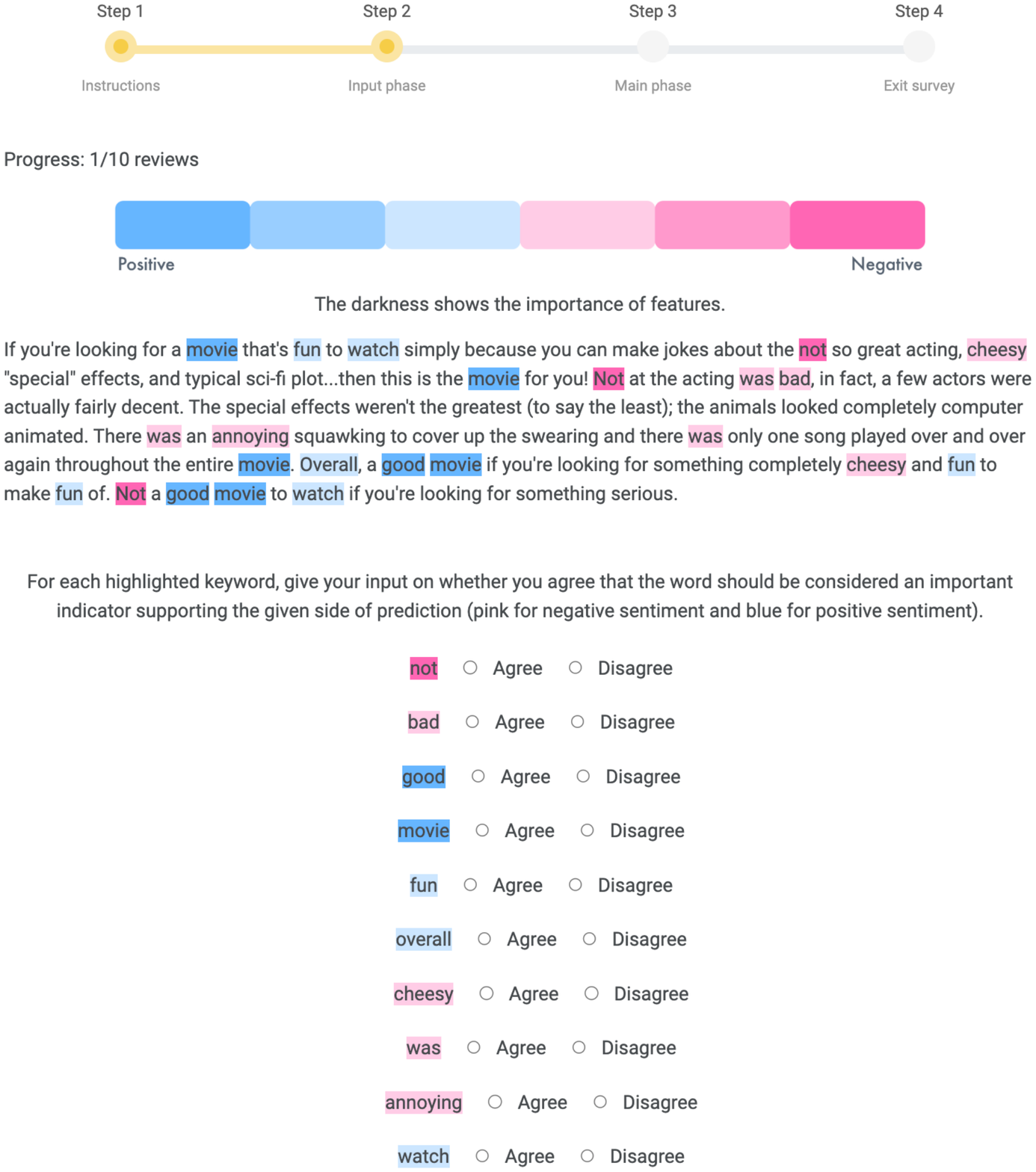}
    \caption{There are two ways of providing input. This figure shows the interface of selective explanations with model explanation critiques (Critique-based).}
    \label{fig:user_study_flow_input_selective}
\end{figure*}

\begin{figure*}[t]
    \centering
    \includegraphics[width=0.95\textwidth]{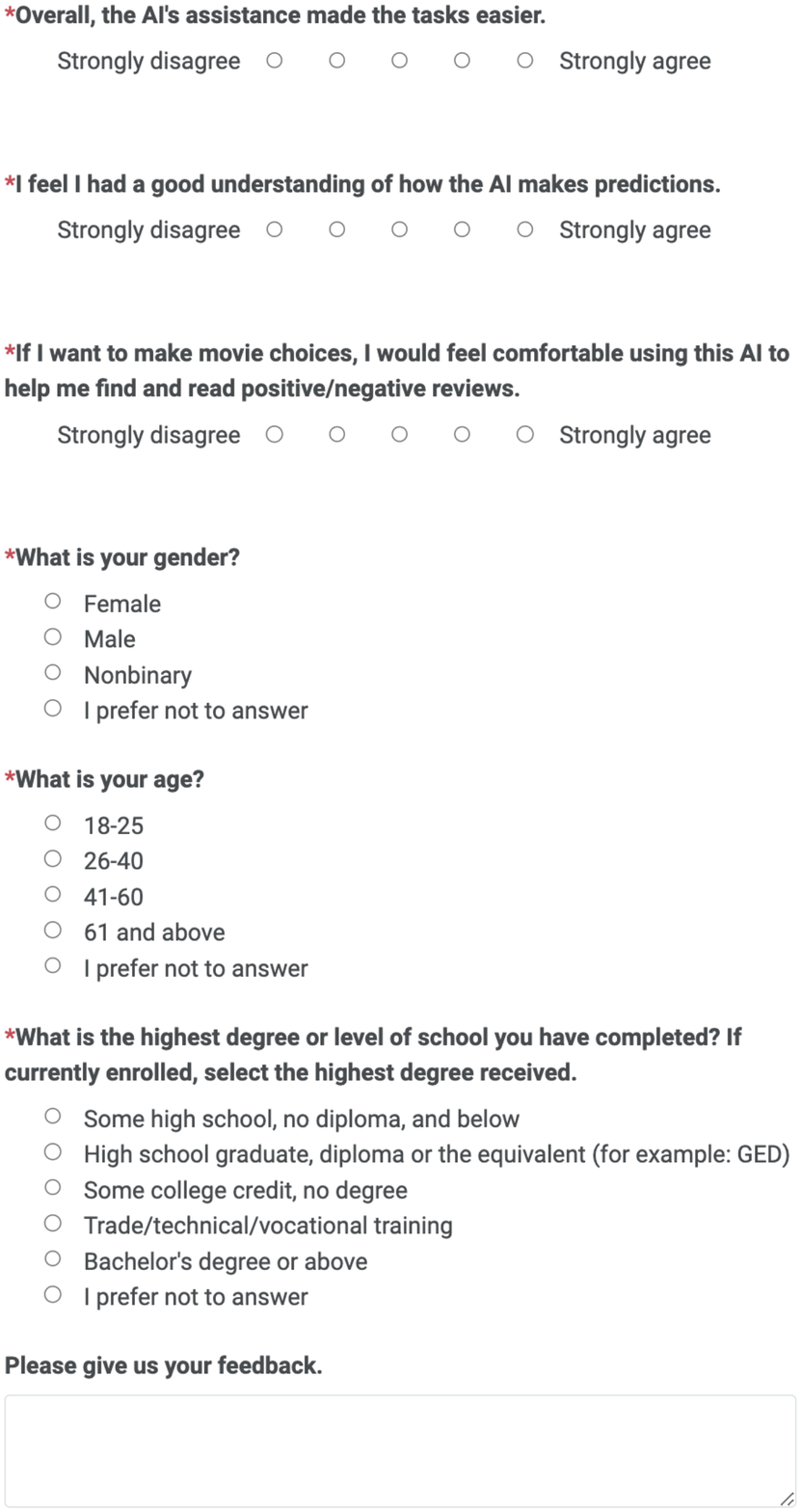}
    \caption{This figure shows the second and remaining part of the survey which features more subjective questions and demographic questions.}
    \label{fig:user_study_flow_survey2}
\end{figure*}

\begin{enumerate}
  \item Read the consent form (see \figref{fig:user_study_flow_consent}), read the instructions, and answer attention check questions (see \figref{fig:user_study_flow_attention1} and \figref{fig:user_study_flow_attention2}).
  
  \item Complete the input phase (see \figref{fig:user_study_flow_input_open_ended} and \figref{fig:user_study_flow_input_selective}). Note that this phase only applies to participants tasked to provide input on their beliefs in explanations.
  
  \item Complete the prediction phase (see \figref{fig:user_study_flow_prediction}).

  \item Complete the task by answering demographic and subjective questions (see \figref{fig:user_study_flow_survey1} and \figref{fig:user_study_flow_survey2}).
\end{enumerate}

\begin{figure*}[t]
    \centering
    \caption{The figure is omitted to preserve anonymity. A participant reads the consent form before commencing the task. The agree button is omitted in the screenshot due to space constraints. Given that the consent form includes the University, we will display the consent form when the submission does not require anonymity.}
    \label{fig:user_study_flow_consent}
\end{figure*}

\begin{figure*}[t]
    \centering
    \includegraphics[angle=-90,width=0.95\textwidth]{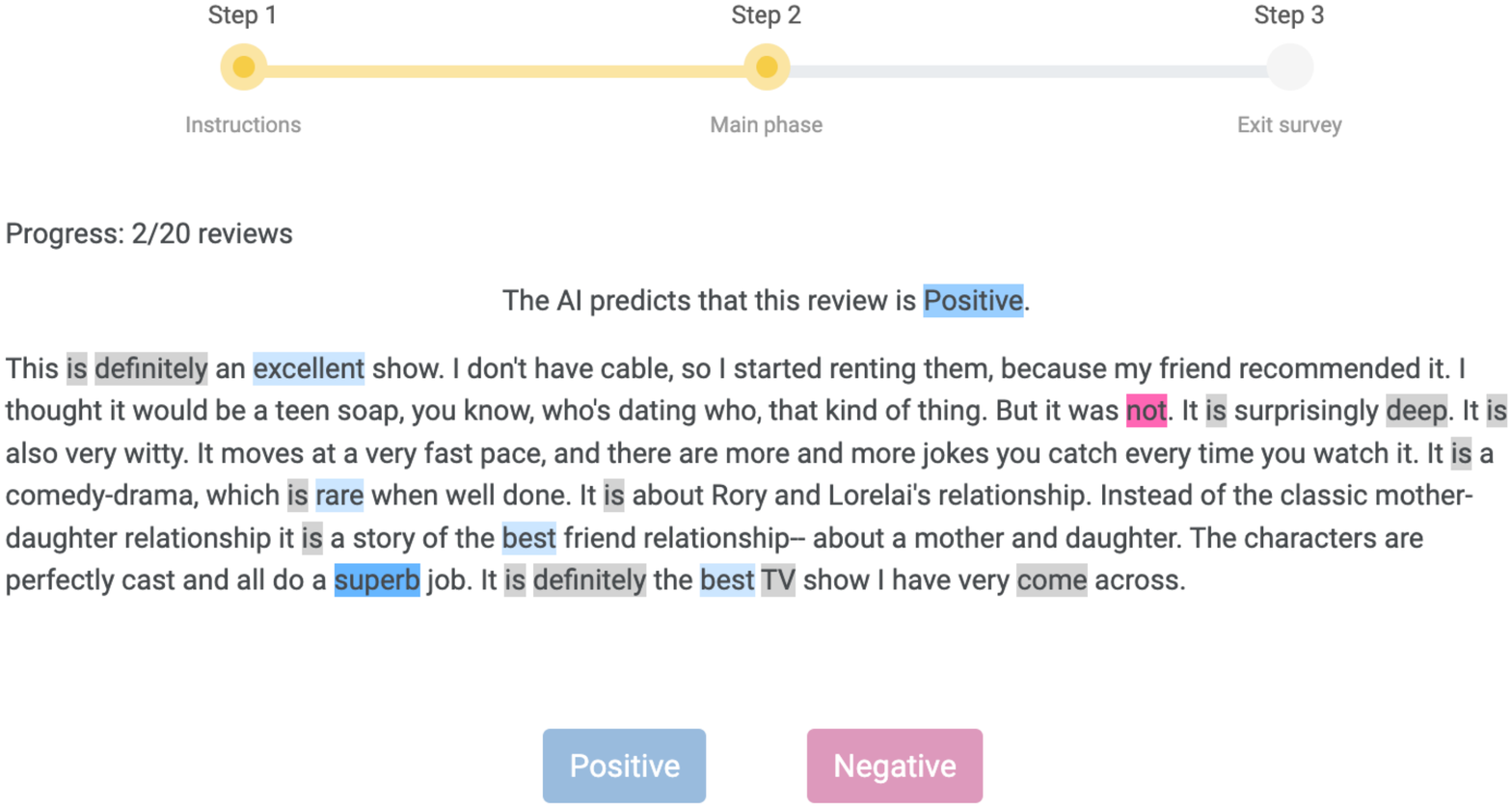}
    \caption{This figure shows the interface of the prediction task. Selective explanations are grayed out and are predicted to be misaligned with what the user would consider as relevant for judging review sentiment.}
    \label{fig:user_study_flow_prediction}
\end{figure*}

\begin{figure*}[t]
    \centering
    \includegraphics[angle=-90,width=0.95\textwidth]{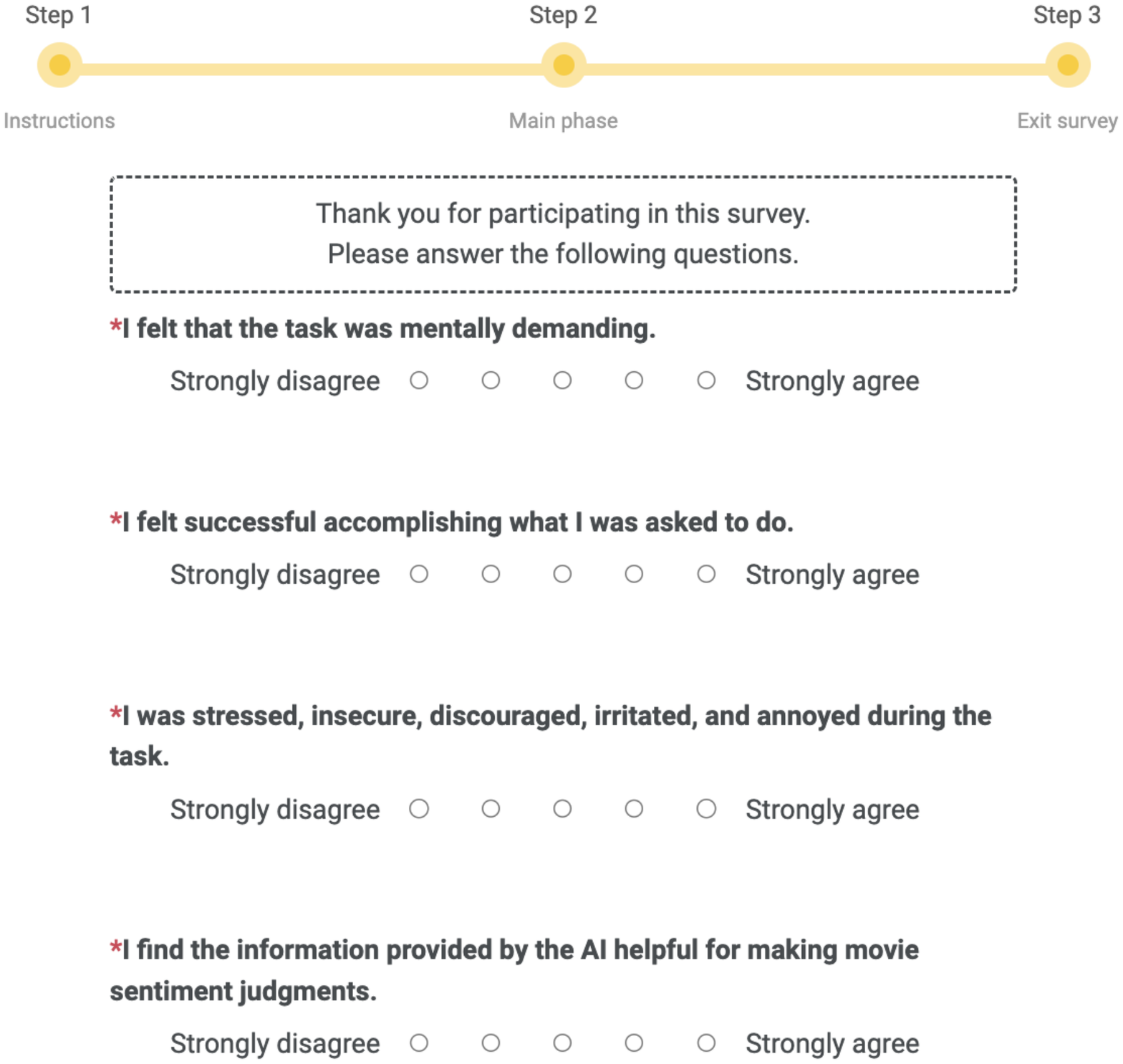}
    \caption{Due to length constraints, this figure shows the first part of the survey which features some of the subjective questions.}
    \label{fig:user_study_flow_prediction}
\end{figure*}

